\UseRawInputEncoding

\documentclass[review,12pt]{elsarticle}

\usepackage[margin=1in]{geometry} 
\usepackage{setspace}             
\usepackage{parskip}              
\usepackage{amssymb}
\usepackage{amsmath}
\usepackage{bm}
\usepackage{booktabs}
\usepackage{multirow}
\usepackage{hyperref}

\usepackage{graphicx}
\usepackage{subcaption}
\usepackage{soul}
\usepackage{float}

\journal{Artificial Intelligence in Medicine}

\usepackage{enumitem}
\usepackage{pifont} 
\usepackage{xcolor} 

\allowdisplaybreaks
  
\begin{document}

\begin{frontmatter}

\title{Robust Rigid and Non-Rigid Medical Image Registration Using Learnable Edge Kernels}

 \author[label1]{Ahsan Raza Siyal}
 \author[label1]{Markus Haltmeier}
 \author[label2,label3]{Ruth Steiger}
 \author[label2,label3]{Malik Galijasevic}
 \author[label2,label3]{Elke Ruth Gizewski}
 \author[label2,label3]{Astrid Ellen Grams}

 \affiliation[label1]{organization={Department of Mathematics, University of Innsbruck},
             country={Austria}}

 \affiliation[label2]{organization={Department of Radiology, Medical University of Innsbruck},
             country={Austria}}

 \affiliation[label3]{organization={Neuroimaging Research Core Facility, Medical University of Innsbruck},
             country={Austria}}



\begin{abstract}
Medical image registration is crucial for various clinical and research applications including disease diagnosis or treatment planning which require alignment of images from different modalities, time points, or subjects. Traditional registration techniques often struggle with challenges such as contrast differences, spatial distortions, and modality-specific variations. To address these limitations, we propose a method that integrates learnable edge kernels with learning-based rigid and non-rigid registration techniques. Unlike conventional layers that learn all features without specific bias, our approach begins with a predefined edge detection kernel, which is then perturbed with random noise. These kernels are learned during training to extract optimal edge features tailored to the task. This adaptive edge detection enhances the registration process by capturing diverse structural features critical in medical imaging. To provide clearer insight into the contribution of each component in our design, we introduce four variant models for rigid registration and four variant models for non-rigid registration. We evaluated our approach using a dataset provided by the Medical University across three setups: rigid registration without skull removal, with skull removal, and non-rigid registration. Additionally, we assessed performance on two publicly available datasets. Across all experiments, our method consistently outperformed state-of-the-art techniques, demonstrating its potential to improve multi-modal image alignment and anatomical structure analysis.
\end{abstract}



\end{frontmatter}

\section{Introduction}
Image registration has been a pivotal area in medical imaging, playing a crucial role in preoperative planning, intraoperative data fusion of different imaging modalities for diagnosis and image guided therapy ~\cite{fu2020deep}. This technique addresses the challenges of rigid and non-rigid deformation by transforming the moving image to align anatomically with the fixed image. At its core, the registration process is formulated as an optimization problem that balances similarity measurements between the fixed and moving images while incorporating a regularization term to prevent irregular deformations. However, conventional methods require re-optimizing for each image pair, leading to high computational costs and slow performance in practical applications. Various iterative optimization techniques ~\cite{Avants2008,Vercauteren2009,Rueckert1999,Thorley2021} have been proposed, many of which have shown excellent accuracy. However, these techniques are often limited by the need for manual adjustments and slow inference speeds.

In recent years, learning-based registration methods have emerged as a powerful alternative, offering fast inference and improved adaptability by leveraging deep neural networks trained across large image datasets. Building on this trend, we introduce an approach that encourages the registration process to focus more directly on the anatomical structures that matter most. Rather than relying solely on global intensity information which can vary significantly across modalities, we incorporate learnable mechanisms that emphasize edge and boundary features, guiding both the rigid and non-rigid alignment stages. This allows the model to better capture structural correspondences, even in the presence of contrast variations or complex tissue deformations. By combining this edge-aware strategy with a two-step registration pipeline, the proposed method aims to deliver more robust and anatomically consistent alignments while maintaining the efficiency of modern learning-based frameworks.

\subsection{Main contributions of our work}

In this work, we propose a method for image registration that integrates learnable edge kernels with learning-based rigid and non-rigid registration techniques.  Learnable edge kernels are designed to emphasize relevant features and anatomical boundaries in medical images, improving alignment across different modalities. By forcing the neural network to focus primarily on edge information, these kernels enhance the accuracy of anatomical structure registration. In the rigid registration phase, learnable edge kernels help manage variations in orientation and scale by concentrating on consistent edge features, ensuring precise alignment despite positional differences. For non-rigid registration, these kernels adapt to local deformations and variations in tissue shape and size, enabling accurate alignment of images with complex anatomical changes.

Incorporating learnable edge kernels into the U-Net architecture can significantly improve registration performance compared to a standard U-Net. Traditional U-Net models process entire image regions, which may include irrelevant or misleading information across different modalities. By integrating learnable edge kernels, the network selectively emphasizes edge features, which are more relevant for registration tasks. This targeted approach enhances the model's ability to capture and align critical anatomical structures, leading to more accurate and reliable registration results. The proposed registration method processes moving and fixed 3D image volumes in two stages. First, a neural network predicts affine transformation parameters, which are then applied to the moving image. Next, another neural network predicts the non-rigid deformation field, using the affine-aligned moving image and the fixed image as inputs. This dual-step approach, enhanced by learnable edge kernels, ensures that the fused images retain essential structural information, thereby improving the overall quality of multimodal imaging analyses. Table~\ref{tab:cnn_vs_edge_kernel} summarizes the key differences between traditional CNN layers and the proposed learnable edge kernels.

\begin{table}[thb!]
    \centering
    \small
    \renewcommand{\arraystretch}{1.25}
    \begin{tabular}{|p{0.22\textwidth}|p{0.36\textwidth}|p{0.36\textwidth}|}
        \hline
        \textbf{Feature} & \textbf{Traditional CNN Layers} & \textbf{Learnable Edge Kernel} \\
        \hline
        Initialization & Random weight initialization (e.g., Gaussian or Xavier) & Predefined with a Laplacian edge kernel as the starting filter \\
        \hline
        Feature Learning & Learns edges together with textures, shapes, and global structures & Starts with edge-specific features and refines them using small perturbations during training \\
        \hline
        Filter Diversity & Filters evolve freely based on backpropagation and gradient updates & Diversity is controlled via random perturbations applied to edge kernels \\
        \hline
        Generalization & Effective across various vision tasks (detection, classification, segmentation) & Specifically optimized for edge detection and structural consistency in registration \\
        \hline
        Robustness to Noise & May struggle with low-contrast images and noise without additional preprocessing & Edge-enhanced initialization improves robustness to noise and low contrast \\
        \hline
        Adaptability in Registration Tasks & Learns alignment features from scratch and may require more data and training time & Naturally adapts to both rigid and non-rigid transformations using edge-focused features \\
        \hline
        Filter Redundancy & Some filters may become redundant or less informative due to unconstrained learning & Controlled diversity encourages each filter to contribute uniquely to edge detection \\
        \hline
        Sensitivity to Small Deformations & May miss fine, small-scale deformations unless explicitly trained for them & Naturally captures local variations in shape, benefiting non-rigid registration \\
        \hline
        Multi-Modal Alignment Performance & Alignment accuracy strongly depends on dataset size and training strategy & Strong edge priors improve alignment efficiency and robustness across modalities \\
        \hline
    \end{tabular}
    \caption{Comparison between traditional CNN layers and the proposed learnable edge kernels.}
    \label{tab:cnn_vs_edge_kernel}
\end{table}

\section{Related Work}

Classical medical image registration has been extensively studied using variational formulations, diffeomorphic mappings, and free-form deformations~\cite{modersitzki2003numerical,sotiras2013deformable,Beg2005,Avants2008,Rueckert1999,Vercauteren2009,Modat2010}. These methods formulate registration as an optimization problem balancing image similarity (e.g., mutual information) and regularization~\cite{Viola1997,Rueckert1999}, and can achieve high accuracy, especially with diffeomorphic models such as LDDMM and SyN~\cite{Beg2005,Avants2008}. However, they require time-consuming per-pair optimization, careful parameter tuning, and remain sensitive to contrast changes and modality-specific variations~\cite{sotiras2013deformable}.

Deep learning has emerged as a powerful alternative that amortizes the registration computation over large datasets~\cite{fu2020deep,duan2025unsupervised}. VoxelMorph and related CNN-based frameworks~\cite{Balakrishnan2018,Balakrishnan2019,Dalca2019} learn a direct mapping from image pairs to deformation fields, greatly reducing runtime while maintaining competitive accuracy. Subsequent unsupervised approaches introduced probabilistic diffeomorphic models~\cite{Dalca2019}, cycle consistency~\cite{Kim2021}, multiscale and adversarial learning~\cite{Lei2020}, variational networks~\cite{Jia2022}, B-spline parameterizations~\cite{Qiu2022}, cascaded schemes~\cite{zhao}, and hyperparameter amortization~\cite{hoopes2021hypermorph}. Adaptive regularization strategies such as DARE~\cite{siyal2025dare} further improve deformation plausibility. Nevertheless, most of these methods remain predominantly intensity-driven, relying on MSE, NCC, or MI-like losses~\cite{Balakrishnan2019,Dalca2019,Kim2021,Lei2020,Qiu2022}, and do not explicitly encode anatomical edge information, which can limit robustness in multimodal or low-contrast scenarios.

Recent work has explored incorporating stronger structure and long-range context into registration architectures. Diffeomorphic and topology-preserving CNN models~\cite{Dalca2019,tony,Qiu2022,Zhang2021,zhou2023self} improve invertibility and smoothness, while preserving anatomical plausibility. Transformer-based registration, including ViT-VNet and TransMorph~\cite{chen2021vit,chen2022transmorph}, leverages global self-attention~\cite{vaswani2017attention,dosovitskiy2020image,liu2021swin} to capture long-range correspondences and has achieved state-of-the-art performance on several benchmarks~\cite{chen2022transmorph,Zhang2021,wang2023modet}. At the same time, U-Net–style and nnU-Net–inspired backbones~\cite{Ronneberger2015,Isensee2020} remain the default design for many registration networks, and recent large-kernel or re-parameterized convolutions~\cite{ding,repvgg} suggest that more structured kernels can improve shape sensitivity. However, these architectures generally treat all features homogeneously and lack explicit, learnable mechanisms that bias the network toward edges and anatomical boundaries.

Across both classical and learning-based methods, rigid and non-rigid alignment are often optimized jointly or without explicit structural guidance~\cite{Balakrishnan2019,Jia2022,Qiu2022}, allowing errors in global alignment and modality-specific appearance to propagate into the deformable stage. This motivates our work: we introduce learnable edge kernels that explicitly bias early feature extraction toward anatomical boundaries and integrate them into separate rigid and non-rigid registration networks. In the next section, we describe our edge-aware design and the four rigid and four non-rigid model variants that systematically evaluate the contribution of these modules to registration performance.

\section{Methods}

We propose an unsupervised, end-to-end 3D deformable neural network specifically designed for brain magnetic resonance image (MRI) registration. Our network is trained in an unsupervised manner by assessing both global and local similarities between image pairs, eliminating the need for ground-truth deformation fields or corresponding anatomical labels. The registration process begins with an affine alignment subnetwork, which predicts the affine transformation required to align the images. Following this, a deformable subnetwork performs non-rigid registration, enabling the accurate alignment of more complex anatomical structures. In this study, we propose four architecture variants for both rigid and non-rigid registration. These variations help to understand the importance of each fundamental component of the architecture.

Let \( f \colon \Omega \to \mathbb{R} \) and \( \bar{m}: \Omega \to \mathbb{R} \) represent the fixed and moving 3D volumes, respectively, both defined on \( \Omega \subseteq \mathbb{R}^3 \). As part of the preprocessing, we first affine-align the images with a network \( G_\theta \) described  using \( m = G_\theta(\bar{m}, f) \). Following this, the images are non-rigidly aligned.

As the main step, we use learned non-rigid registration for aligning \( f \) and \( m \). Instead of separately determining a deformation field \( \Phi = \text{Id} + u \) for each pair \( (f, m) \), the learning approach takes \( u = U_\theta(f, m) \), where \( U_\theta \) is a parametric function, such as a convolutional neural network (CNN), with \( \theta \) as its learnable parameters. If ground truth data \( (f, m, u) \) were available, the network could be trained to minimize the supervised loss based targeting \(  U_\theta(f, m) \simeq  u \). Due to the lack of ground truth displacement fields in most cases, we use an unsupervised approach. Here, the CNN \( U_\theta \) is selected by minimizing the loss. 
\[
L(\theta) \triangleq \sum_{t =1}^T  D(f_t, m_t \circ \Phi) + \alpha R(\Phi) \,,
\]
where \( \{ (f_t, m_t) \colon t =1, \dots, T\} \) are the training data,  \( \alpha >0 \) the regularization parameter,  \( D  \) the data fidelity term  and $R$ the regularizer.

\subsection{Components of the Proposed Architectures}

\subsubsection{Learnable Edge Kernels}

Learnable edge kernels enhance multi-modal medical image alignment by training the network to focus more on edges, thereby highlighting key anatomical features. In rigid registration, it manages orientation and scale variations, while in non-rigid registration, it adapts to local deformations, ensuring accurate alignment and retention of critical structural information. This improves the quality of multi-modal imaging analyses. 
The proposed learnable edge kernel in a neural network context starts with a predefined edge detection function but is trainable, allowing its weights to adapt during training. The convolutional weights are initialized using the Laplacian edge kernel, followed by Leaky ReLU activations. Each weight in the convolutional kernel is slightly adjusted by multiplying it with a factor that includes small random perturbations, ensuring that no two convolutional filters are exactly the same. This variation introduces diversity among channels, enabling the network to learn a broader set of edge features. 

Additionally, use maximum response selection to select the most diverse 16 channels. The idea is to calculate the mean activation of each channel across the spatial dimensions (depth, height, width) and then select the top \( n \) channels with the highest mean activation values. The benefits of this approach include enhanced generalization and robustness in detecting edges across various orientations and contexts within 3D data. This is crucial for applications like medical imaging, where accurate edge detection significantly impacts diagnosis and analysis. If the original weight is \( w_{ijk} \), the new weight becomes  

\[
\hat{w}_{ijk} = w_{ijk} \times (1 + 0.1 \times \mathcal{N}(0, 1))
\]

where \( \mathcal{N}(0, 1) \) represents the standard normal distribution.  

\begin{figure}[thb!]
\centering
\includegraphics[width=0.6\linewidth]{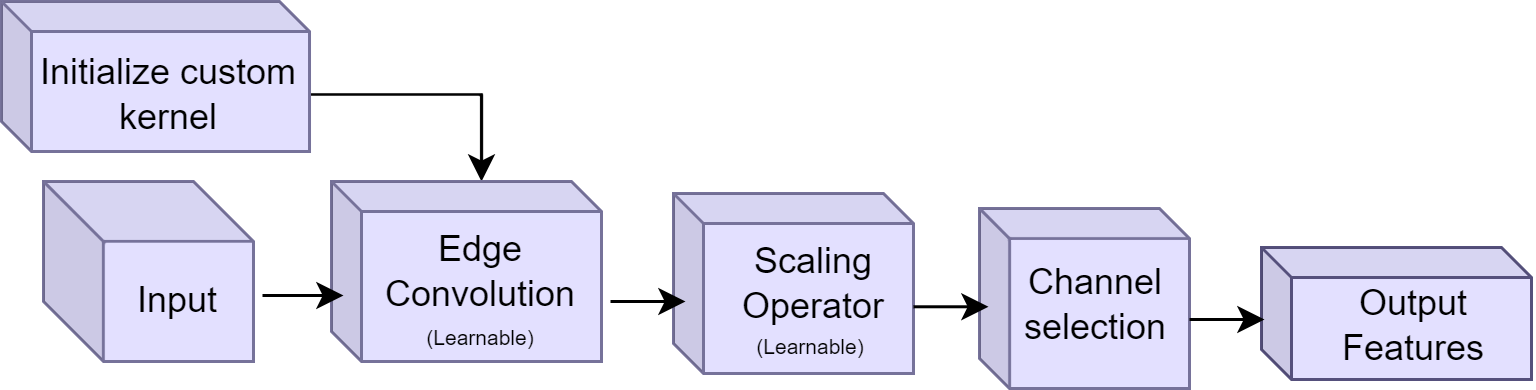}\\[2em]
\includegraphics[width=0.6\linewidth]{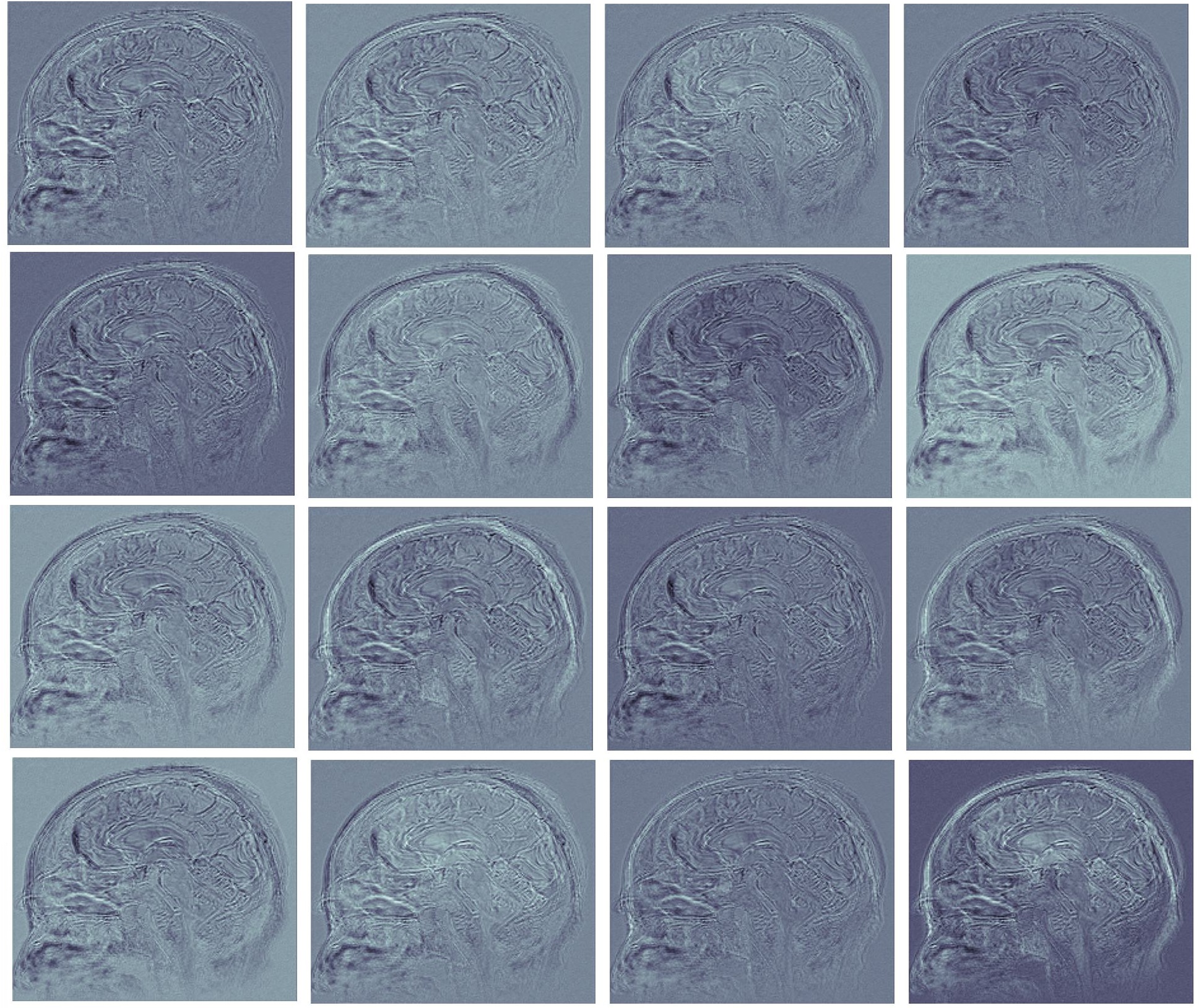}
\caption{Top: Architecture of the learnable edge detection module. Bottom: Features extracted by the learnable edge detection module.}
\label{fig:edge_module}
\end{figure}

Traditional convolutional layers and the learnable edge detection module both aim to learn features from input data but differ significantly in their approach. Traditional layers start with random weights and learn all features, including edges, through data-driven training, making them flexible and general-purpose for various tasks. In contrast, the learnable edge detection module begins with a predefined edge detection kernel, providing a strong prior specifically for edge detection, which is then slightly perturbed with random noise to introduce variation. This approach ensures that each filter is slightly different, enhancing the module's ability to detect a diverse set of edge features.   This specialization makes the learnable edge detection module particularly effective for tasks where edge information is crucial, such as medical imaging, ensuring accurate and robust detection of key anatomical structures. Figure~\ref{fig:edge_module} shows the graphical representation of the learnable edge detection module (top) and the features extracted by the module given a concatenated fixed and moving image (bottom).

\subsubsection{Residual Convolution and Dilated Convolution Blocks}

\begin{figure}[thb!]
     \centering
     \begin{subfigure}[b]{0.35\textwidth}
         \centering
         \includegraphics[width=\textwidth]{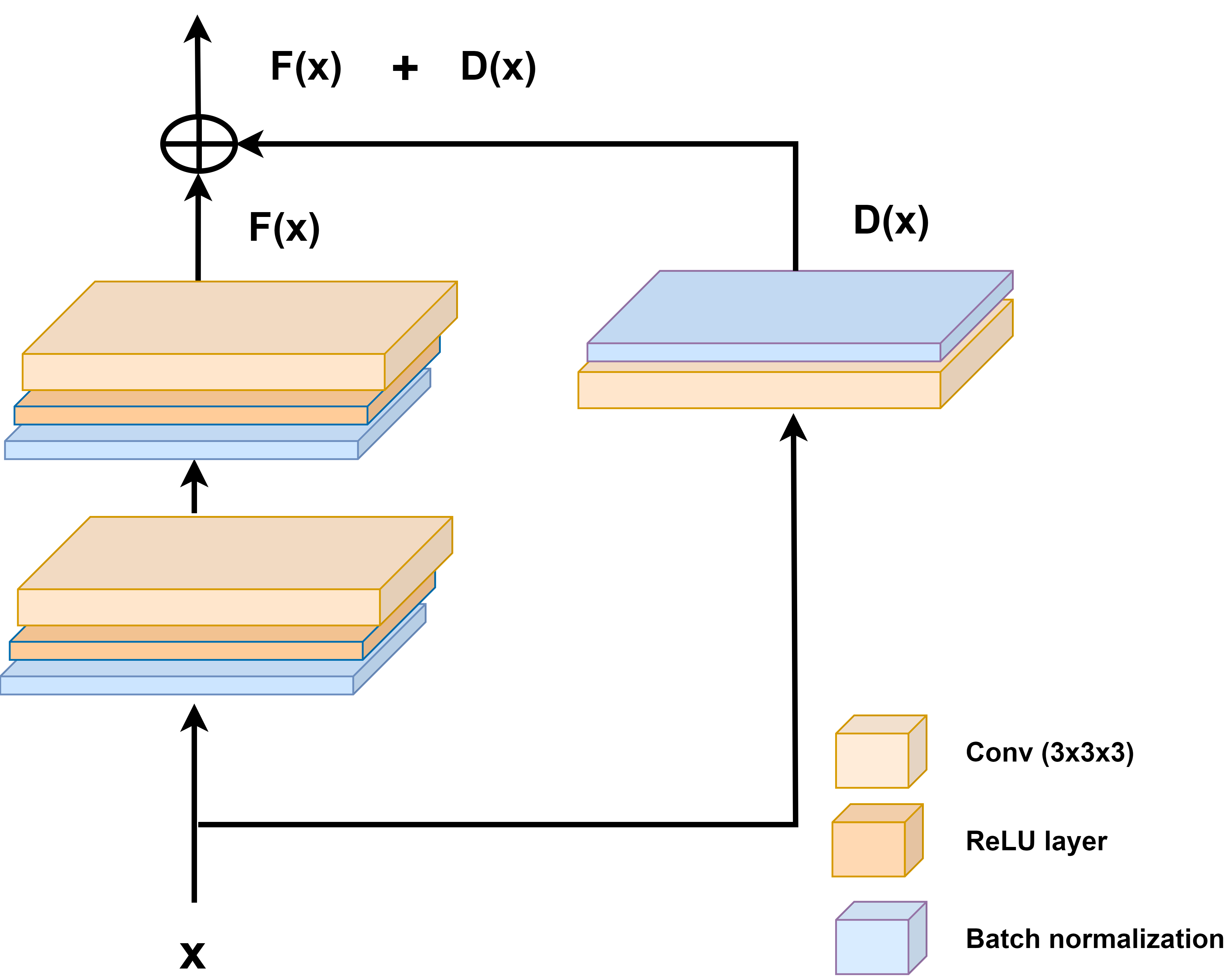}
         \caption{The residual module architecture }
         \label{fig:y equals x}
     \end{subfigure}
     \hfill
     \begin{subfigure}[b]{0.35\textwidth}
         \centering
         \includegraphics[width=\textwidth]{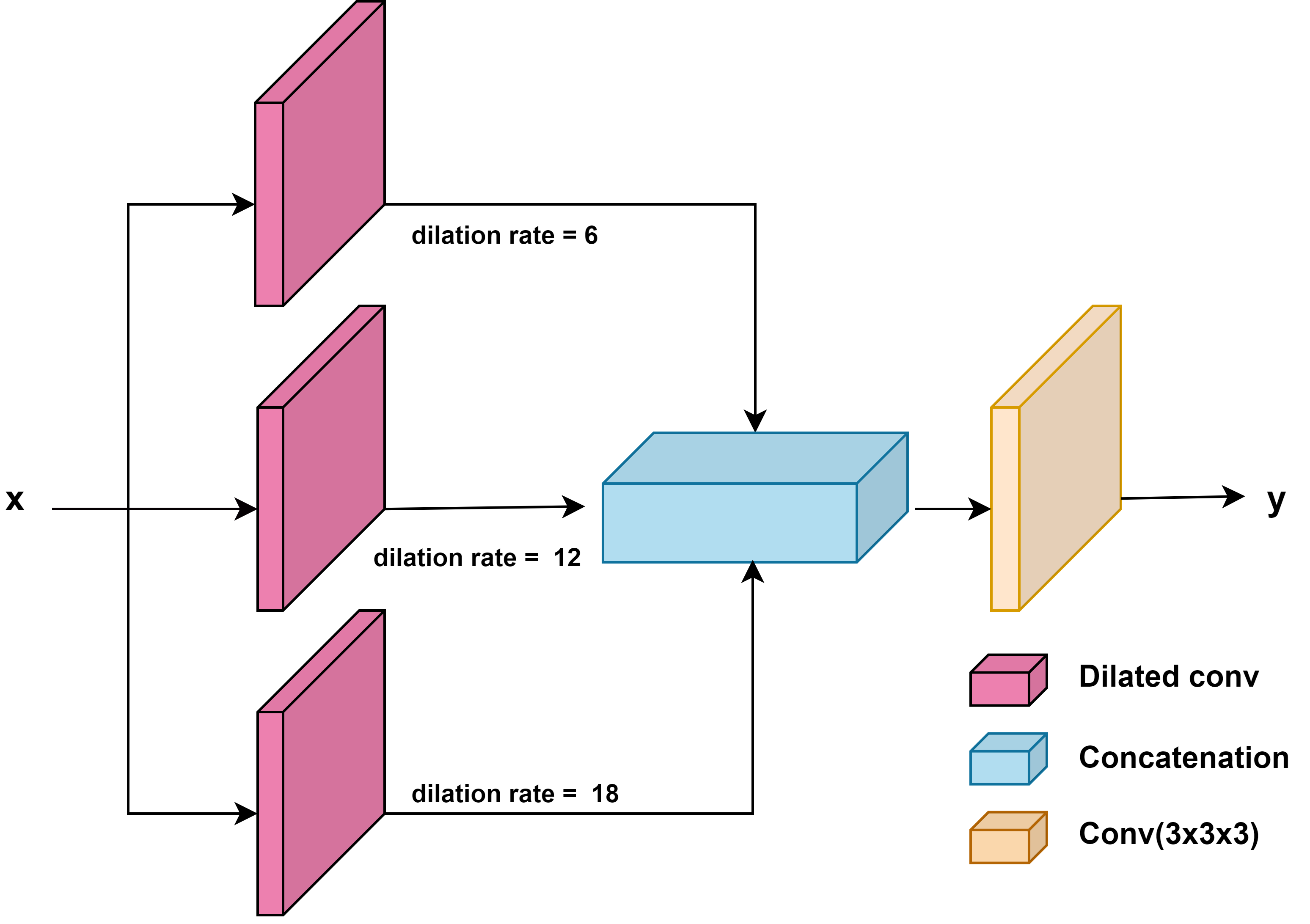}
         \caption{Dilated convolution module architecture}
         \label{fig:three sin x}
     \end{subfigure}
             \caption{Residual Block and Dilated Convolution Block Architectures: The residual block (top) uses convolutional layers with shortcut connections for efficient gradient flow, while the dilated convolution block (bottom) employs multiple dilations for multi-scale contextual information capture.}
        \label{fig:res_dia}
\end{figure}

A 3D convolutional block with residual connections, combining the output of two convolutional layers with a skip connection. It includes batch normalization, ReLU activation, and dropout for regularization. It sums the outputs of the main convolutional block and the skip connection, facilitating gradient flow. 
The dilated convolution module, utilizing multiple convolutional layers with different dilation rates to capture multi-scale features. Each block applies 3D convolutions with specified dilation rates of 6,12 and 18, followed by ReLU activation and batch normalization. The forward method concatenates the outputs of these blocks and passes them through a final convolutional layer for dimensionality reduction. Figure \ref{fig:res_dia} shows the architectures.

\subsubsection{Inception Module}

\begin{figure}[thb!]
\centering
\includegraphics[width=0.6\linewidth]{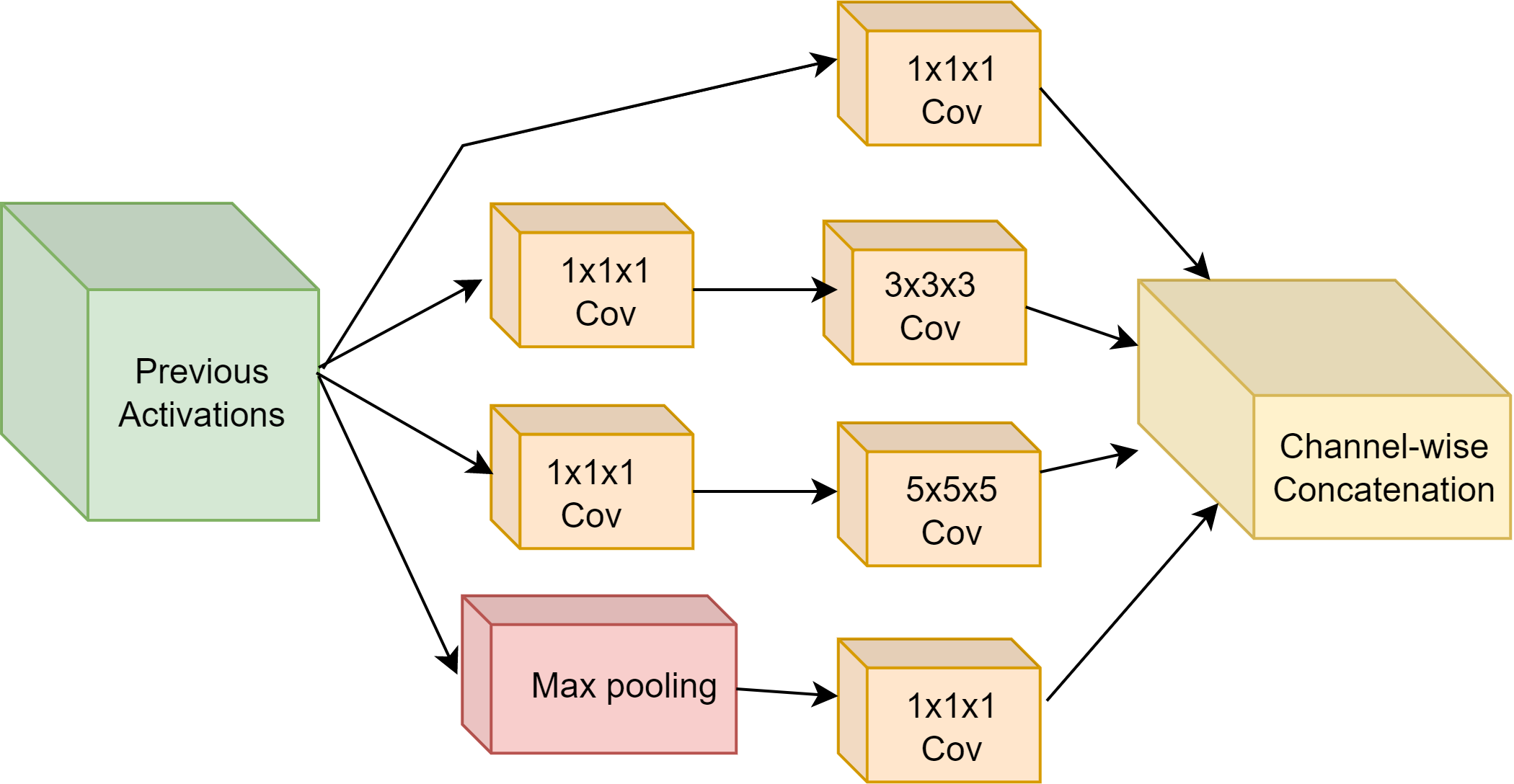}
\caption{The inception module consists of parallel convolutional layers with different filter sizes ($1 \times 1$, $3 \times 3$, and $5 \times 5$), allowing for multi-scale feature extraction. The outputs of these layers are concatenated along the channel dimension.}
\label{fig:incep_module}
\end{figure}

The Inception Module is a multi-branch network module designed to capture features at different scales. It has four parallel branches: a $1 \times 1$ convolution, a $1 \times 1$ followed by $3 \times 3$ convolution, a $1 \times 1$ followed by $5 \times 5$ convolution, and a $3 \times 3$ max pooling followed by a $1 \times 1$ convolution. It concatenates the outputs from all branches along the channel dimension. Figure \ref{fig:incep_module} shows the architecture of Inception module used in this study.

\subsubsection{Dense Feature Fusion Module}

\begin{figure}[thb!]
\centering
\includegraphics[width=0.6\linewidth]{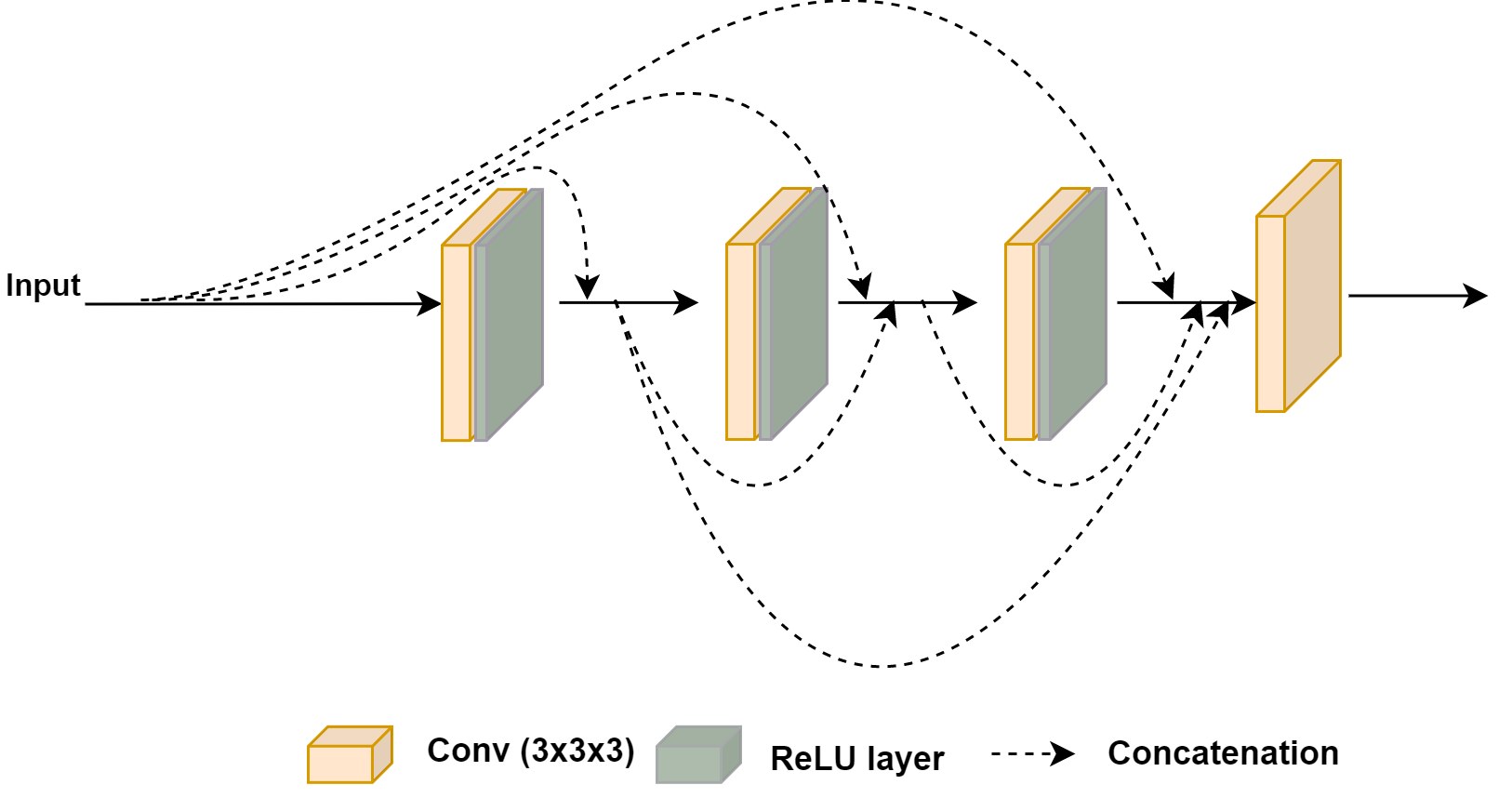}
\caption{The architecture of the Dense Feature Fusion Module.}
\label{fig:dense_module}
\end{figure}

The Dense Feature Fusion Block is a densely connected convolutional block, where the output of each convolution layer is concatenated with its input and passed to the next layer. It has three sequential 3D convolution layers, each producing 16 feature maps, followed by a final convolution that reduces the spatial dimensions. The concatenation of all intermediate outputs allows for richer feature extraction before the final convolution and downsampling. Figure \ref{fig:dense_module} shows the architecture of the module.

\subsection{Rigid Registration}

Affine registration is a critical technique in image processing and computer vision, particularly in medical imaging, where it is used to align two volumetric images: a fixed and a moving volumes. The objective is to find a set of transformation parameters that can map the moving volume to the fixed volume. Affine registration predicts 12 degrees of freedom (DOF) affine parameters, which include three translations, three rotations, three scales, and three shear components.Variants have been introduced in the article to explore different approaches for enhancing feature extraction and alignment accuracy in medical image registration. Each variant builds on the previous one, incorporating distinct modules such as Laplacian filtering and learnable edge detection, to refine the processing of moving and fixed images. These modifications aim to improve the precision of affine parameter regression, crucial for accurate volume alignment.

\subsubsection{Reg-LEdge-Model Variant 1 for rigid registration}

\begin{figure}[thb!]
\centering
\includegraphics[width=1\linewidth]{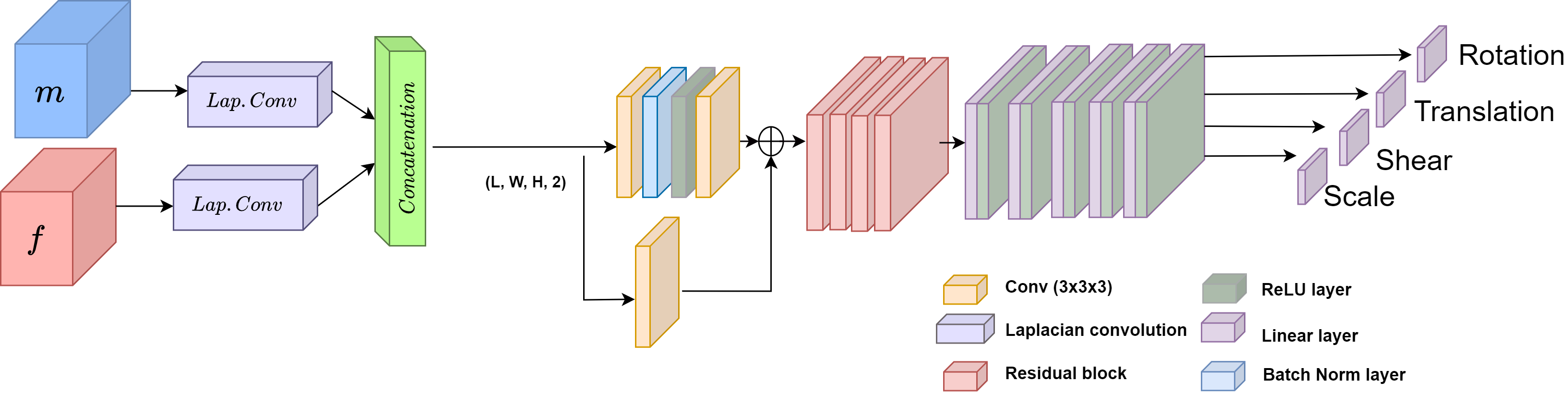}
\caption{Reg-LEdge-Model Variant-1 architecture (rigid): takes Laplacian convoluted moving and fixed volumes and outputs affine parameters}
\label{fig:rigid_variant1}
\end{figure}

In this variant, the Laplacian kernel is individually applied to both the moving and fixed volumes, which are then concatenated. This is followed by multiple residual convolutional blocks to capture and process hierarchical features. Laplacian filtering enhances edge detection, thereby improving feature extraction and alignment accuracy. The upsampling stages refine the feature maps, which are subsequently passed through fully connected layers to regress affine parameters, including rotation, scaling, translation, and shear. These parameters are essential for aligning the moving volume with the fixed volume. Figure \ref{fig:rigid_variant1} illustrates the proposed architecture for Reg-LEdge-Model Variant 1.

\subsubsection{Reg-LEdge-Model Variant 2}
\begin{figure}[thb!]
\centering
\includegraphics[width=0.5\linewidth]{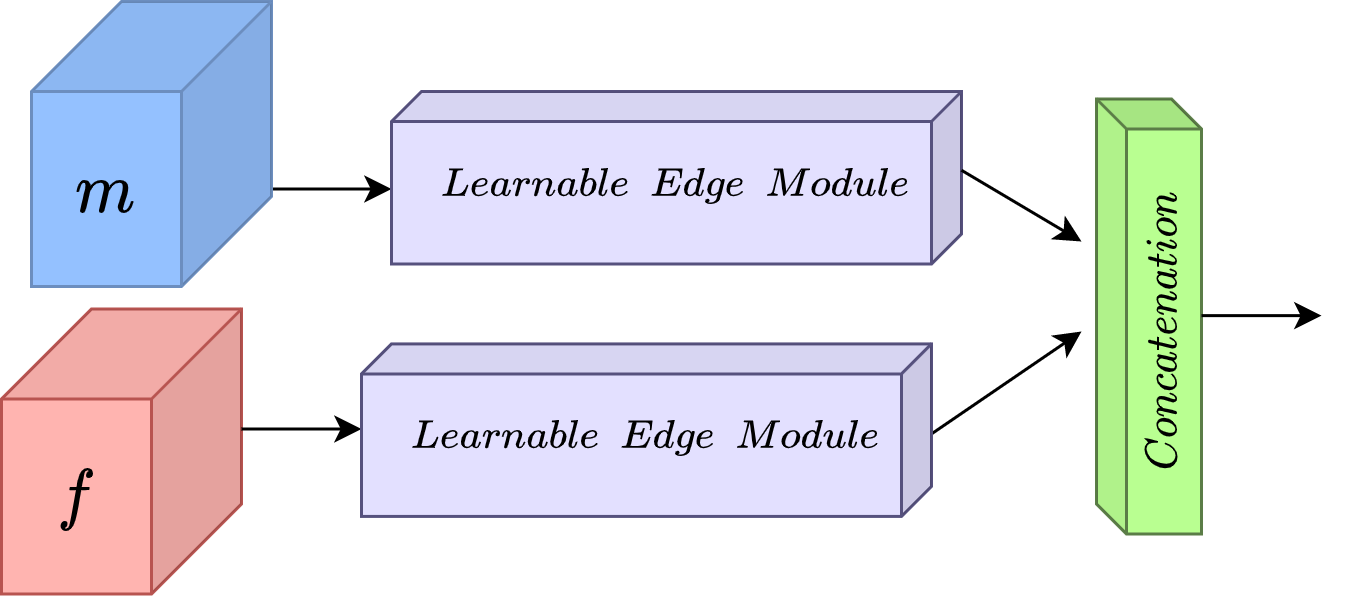}
\caption{Reg-LEdge-Model Variant-2 architecture (rigid): Moving and fixed volumes are passes through learnable edge module and rest of the architecture is similar to Reg-LEdge-Model variant-1}
\label{fig:rigid_variant2}
\end{figure}

 In this variant, both the moving and fixed volumes are individually passed through a learnable edge detection module and later concatenated. The rest of the architecture is similar to Reg-LEdge-Model Variant 1. Figure \ref{fig:rigid_variant2} shows the graphically the architecture.
 
 \subsubsection{Reg-LEdge-Model Variant 3}
\begin{figure}[thb!]
\centering
\includegraphics[width=0.8\linewidth]{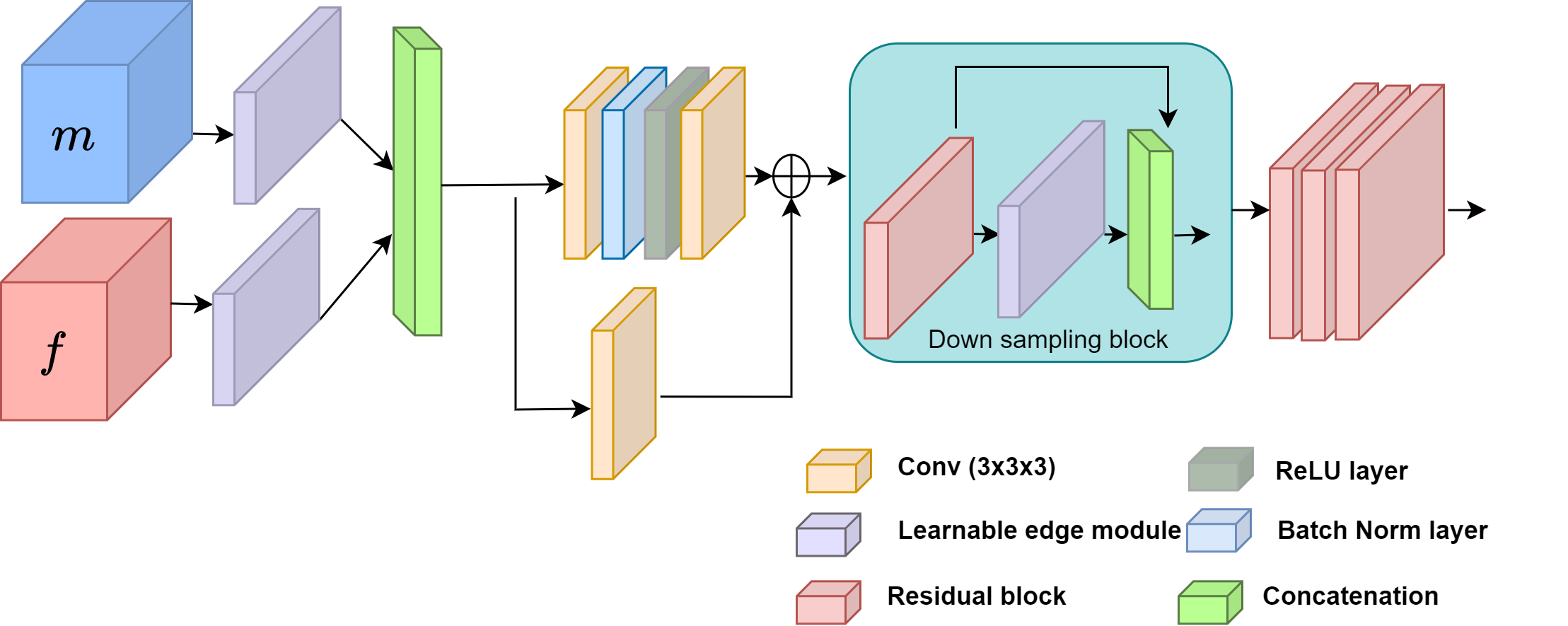}
\caption{Reg-LEdge-Model Variant-3 architecture (rigid): The moving and fixed volumes are first processed through a learnable edge detection module and then concatenated. This combined output is passed through convolution layers with residual connections, followed by a down-sampling block and additional residual convolutional blocks. The decoder part, similar to Reg-LEdge-Model Variant-1, is not shown}
\label{fig:rigid_variant3}
\end{figure}

The moving and fixed volumes are passed through a learnable edge detection module and later concatenated. This is then passed through a set of convolution layers and residual connection. After that, a down-sampling block is applied, followed by consecutive residual convolutional blocks. In this variant we used edge detection module only to first block of down-sampling arm. Therefore, the edges are detected in the early stages and rest is simply residual convolutions. The decoder part is like Reg-LEdge-Model variant 1. Figure \ref{fig:rigid_variant3} shows the proposed architecture of Reg-LEdge-Model variant-3
 
 \subsubsection{Reg-LEdge-Model Variant 4}
 
\begin{figure}[thb!]
\centering
\includegraphics[width=1\linewidth]{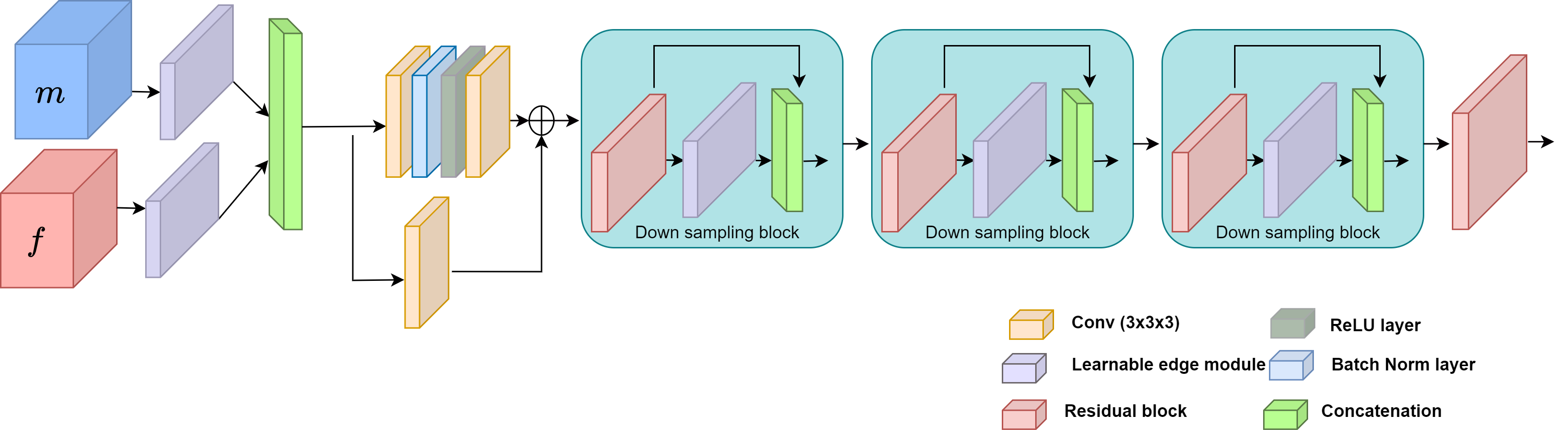}
\caption{Reg-LEdge-Model Variant-4 architecture(rigid): the down-sampling block is used three times, serving exclusively for deep feature extraction through down-sampling.Rest is very similar to Reg-LEdge-Model variant-3}
\label{fig:rigid_variant4}
\end{figure}

In this variant, the moving and fixed volumes are first passed through a learnable edge detection module and then concatenated. This combined output is subsequently passed through a series of convolution layers and residual connections. Following this, three consecutive down-sampling blocks are applied, each of which performs down-sampling while incorporating edge detection within every block of the down-sampling arm. The decoder part of the architecture is similar to that in Reg-LEdge-Model Variant 1. Figure \ref{fig:rigid_variant4} graphically illustrates the architecture for this variant.

 \subsection{Non-Rigid Registration}
 Deformable image registration involves transforming one image to align with another by finding a deformation $\Phi$ that maps one image domain to another, ensuring that the fixed image $f$ aligns with the warped moving image $m$ composed with $\Phi$. This process is mathematically complex due to its inherent ill-posed nature, characterized by non-uniqueness and instability from image noise. To mitigate these issues, regularization techniques are used to incorporate prior information and stabilize the solution. In this study, we propose four model architectural variants to explore different architectural possibilities.
 
 \subsubsection{Reg-LEdge-U-Model Variant 1 for non-rigid registration}
\begin{figure}[thb!]
\centering
\includegraphics[width=0.6\linewidth]{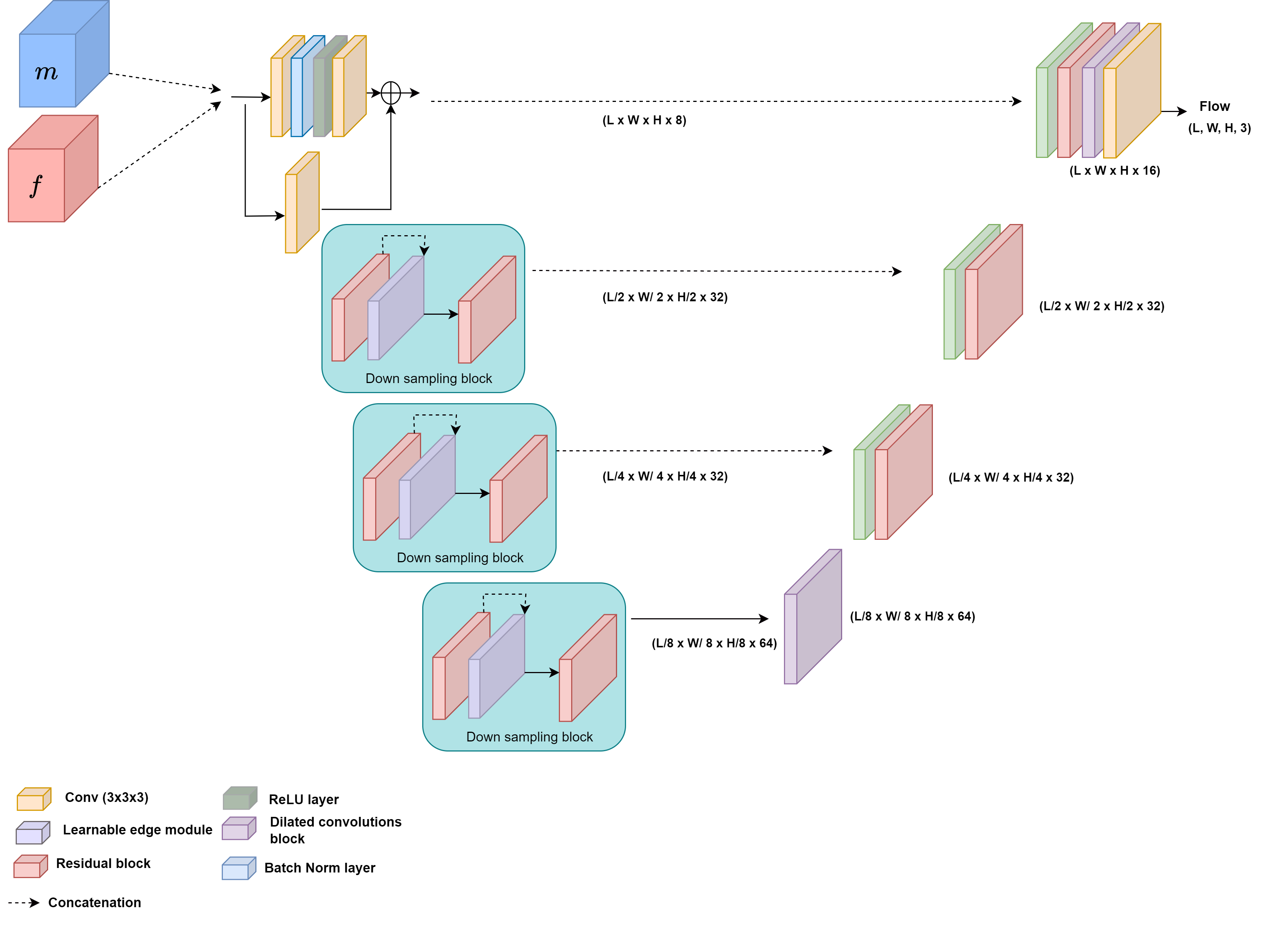}
\caption{Shows the graphical representation of the Reg-LEdge-U-Model Variant-1 architecture (non-rigid)}
\label{fig:non_rigid_variant1}
\end{figure}

In this variant, the model processes a concatenated input of moving and fixed volumes, which is passed through a series of convolution layers with residual connections. Following these layers, three consecutive down-sampling blocks are applied, each consisting of a residual convolutional block and an edge detection module. At the bottleneck, dilated convolution block is utilized to capture and process hierarchical features. The upsampling arm comprises a residual block and ReLU activation, ultimately generating a displacement field as the network's output. Figure \ref{fig:non_rigid_variant1} shows the overall architecture of Reg-LEdge-U-Model variant-1 for non-rigid registration.
 
 \subsubsection{Reg-LEdge-U-Model Variant 2}
\begin{figure}[thb!]
\centering
\includegraphics[width=0.5\linewidth]{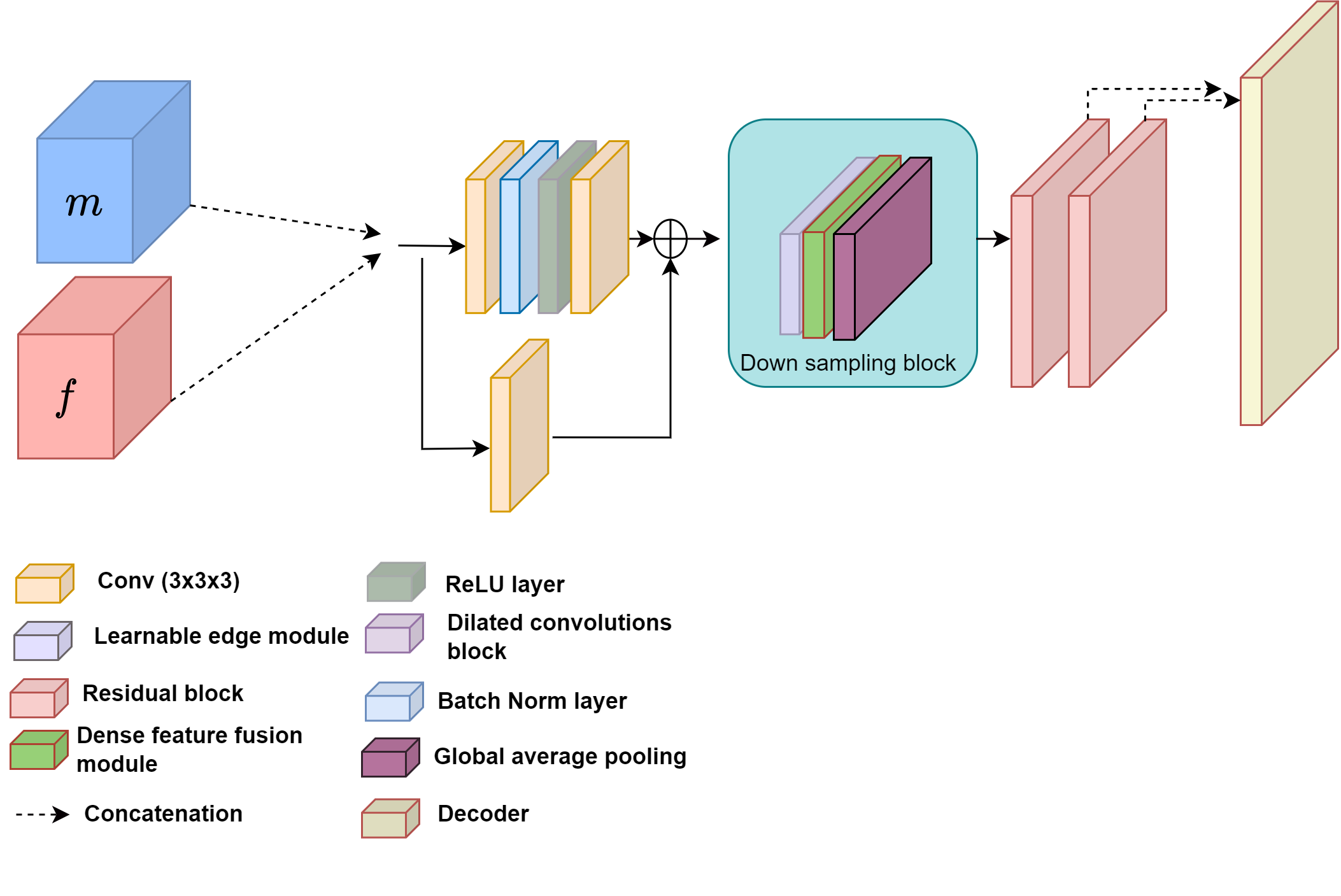}
\caption{Shows the graphical representation of the Reg-LEdge-U-Model Variant-2 architecture(non-rigid)}
\label{fig:non_rigid_variant2}
\end{figure}

In this variant, the model receives a concatenated input of a moving and a fixed volumes. This combined input is processed through a sequence of convolutional layers with residual connections. Following these layers, a down-sampling block is applied, consisting of an edge detection module, dense feature fusion, and average pooling layers, followed by two residual blocks for additional down-sampling. The bottleneck and upsampling components are identical to those in Reg-LEdge-U-Model Variant 1. Figure \ref{fig:non_rigid_variant2} shows graphical representation.
 
 \subsubsection{Reg-LEdge-U-Model Variant 3}
\begin{figure}[thb!]
\centering
\includegraphics[width=0.5\linewidth]{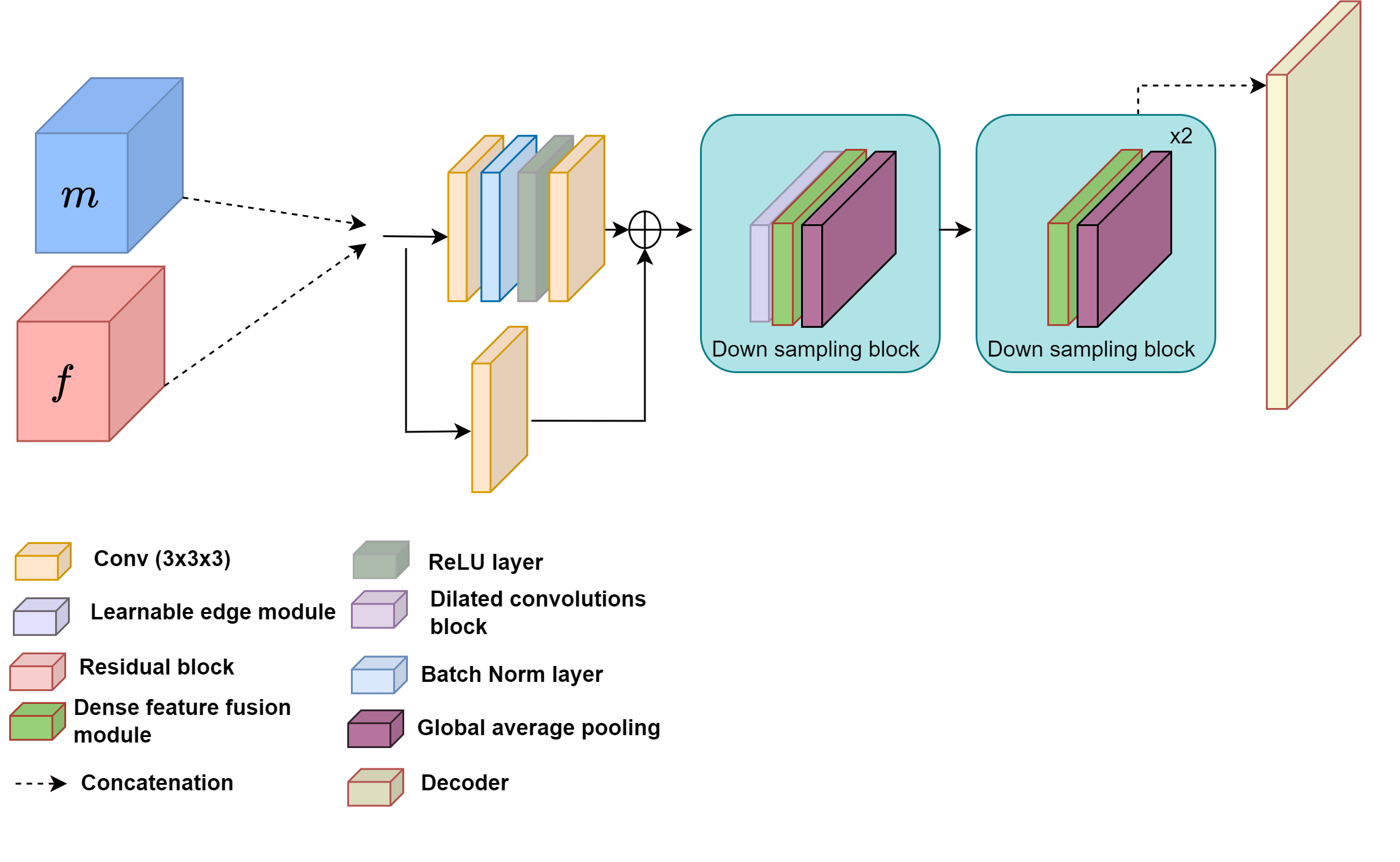}
\caption{Shows the graphical representation of the Reg-LEdge-U-Model Variant-3 architecture(non-rigid)}
\label{fig:non_rigid_variant3}
\end{figure}

In this variant, the model takes a concatenated input of a moving and a fixed volumes. This combined input is processed through a series of convolutional layers with residual connections. Following these layers, a down-sampling block with dense feature fusion is applied. Subsequently, two consecutive dense feature modules with average pooling are used for further down-sampling. The bottleneck and upsampling components are identical to those in Reg-LEdge-U-Model Variant 1. Figure \ref{fig:non_rigid_variant3} shows the architecture.
 
 \subsubsection{Reg-LEdge-U-Model Variant 4}
\begin{figure}[thb!]
\centering
\includegraphics[width=0.6\linewidth]{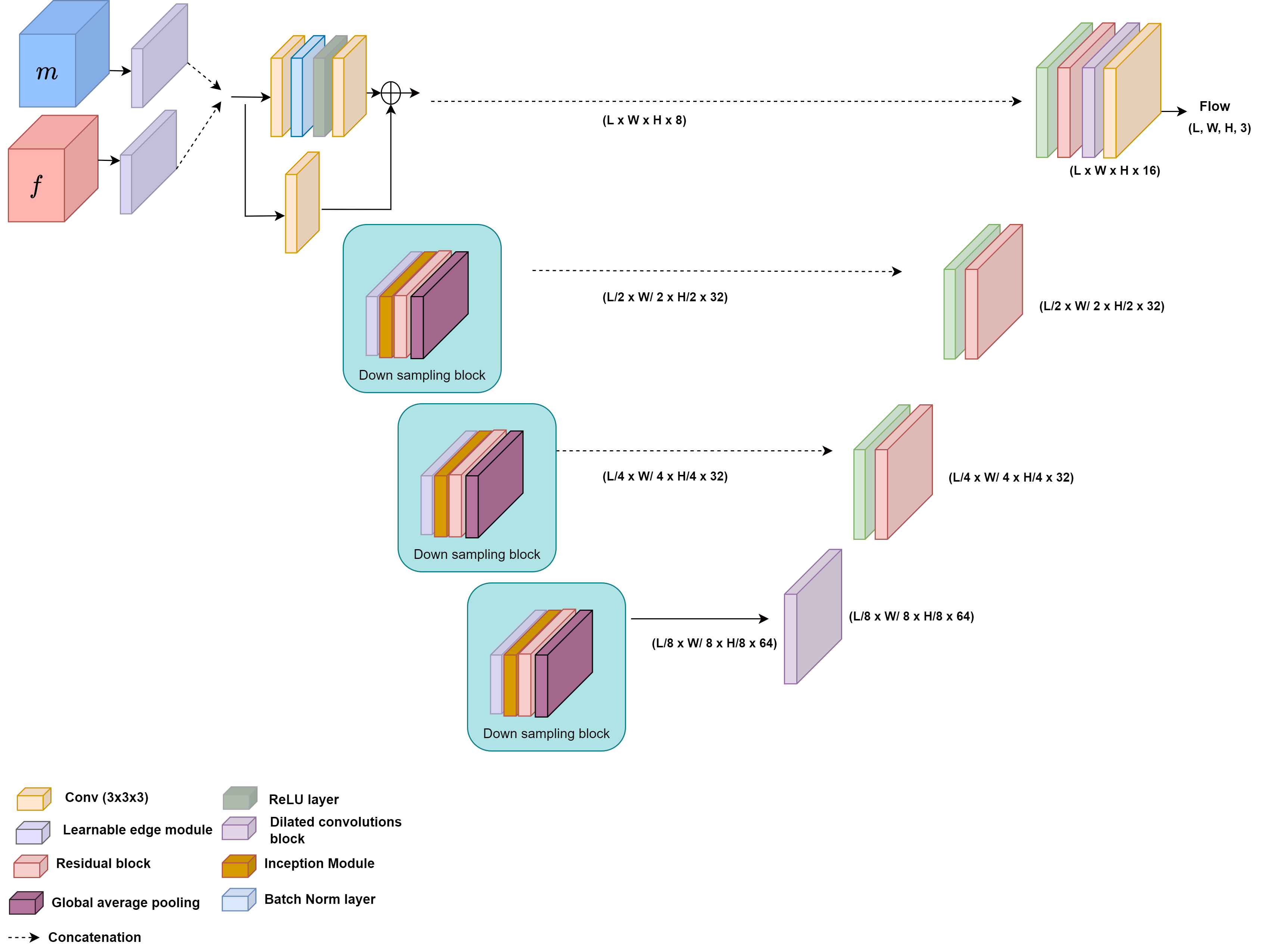}
\caption{Shows the graphical representation of the Reg-LEdge-U-Model Variant-4 architecture(non-rigid)}
\label{fig:non_rigid_variant4}
\end{figure}

 The moving and fixed volumes are first processed through a learnable edge detection module and then concatenated. This combined input is passed through a series of convolutional layers with residual connections. The encoder part employs three consecutive down-sampling blocks. Each down-sampling block consists of a learnable edge detection module, an inception module, a residual convolutional module, and concludes with average pooling. The bottleneck and decoder components remain the same as in Reg-LEdge-U-Model variant 1. Figure \ref{fig:non_rigid_variant4} shows the architecture.
 
 \subsection{Network loss}
 The loss function is composed of two key components: a similarity loss, denoted as $D(m \circ \Phi, f)$, which measures the appearance differences between images, and a smoothness loss, $R(\Phi)$, that penalizes spatial irregularities in the displacement field. To evaluate the similarity between a fixed image $f$ and a warped image $W = m \circ \Phi$, we utilize mutual information $\mathcal{D}_{\mathrm{MI}}(f, W)$, which is particularly robust to intensity variations, commonly encountered in medical imaging data. The regularization term $R(\Phi)$ is given by:

\[
R(\Phi) = \sum_{p \in \Omega} \|\nabla \Phi(p)\|^2 \,.
\]

Incorporating local mutual information, the similarity measure can be expressed as:

\[
\mathcal{D}_{\mathrm{LMI}}(f, W) = -\sum_{p \in \Omega} \sum_{i,j} p_{fW}(i,j,p) \log\left(\frac{p_{fW}(i,j,p)}{p_f(i,p) p_W(j,p)}\right) \,,
\]

where $p_{fW}(i,j,p)$ represents the joint probability distribution of intensities at each point $p$, and $p_f(i,p)$ and $p_W(j,p)$ are the marginal probability distributions of the intensities in the fixed and warped images, respectively.
\subsection{Spatial transformation functions}
\subsubsection{Affine transformation}
In 3D space, a 12-degree-of-freedom (12-DOF) affine transformation is a linear transformation followed by a translation. It can include translation, rotation, scaling, and shearing operations. The transformation is applied to a point in homogeneous coordinates. The affine transformation in 3D space can be expressed as:

$$y = Ax + b$$

where, $x$ is the original point in 3D space, $y$ is the transformed point, $A$  is a $ 3 \times 3 $ matrix representing the linear part of the transformation and $b$ is the translation vector.

In homogeneous coordinates, this transformation can be written as:
\[
\begin{pmatrix}
y_1 \\
y_2 \\
y_3 \\
1
\end{pmatrix}
=
\begin{pmatrix}
a_{11} & a_{12} & a_{13} & b_1 \\
a_{21} & a_{22} & a_{23} & b_2 \\
a_{31} & a_{32} & a_{33} & b_3 \\
0 & 0 & 0 & 1
\end{pmatrix}
\begin{pmatrix}
x_1 \\
x_2 \\
x_3 \\
1
\end{pmatrix}
\]

where:
\begin{itemize}
    \item \( a_{11}, a_{22}, a_{33} \) control scaling along the \( x \), \( y \), and \( z \) axes, respectively.
    \item \( a_{12}, a_{13}, a_{21}, a_{23}, a_{31}, a_{32} \) represent shearing factors, which skew the axes relative to one another.
    \item \( b_1, b_2, b_3 \) control translation along the \( x \), \( y \), and \( z \) axes, respectively.
\end{itemize}
Here, $A$ is now represented as a \( 3 \times 3 \) matrix embedded within a \( 4 \times 4 \) matrix, along with the translation vector $b$.

\subsubsection{Non-rigid spatial transformation}
We use a differentiable operation based on spatial transformer networks to calculate $M \circ \phi$. We calculate a voxel position $\phi (p)$ in
$M$ for each voxel $p$. The values at the eight nearby voxels are linearly interpolated because image values can only be specified at integer locations.
\begin{equation}
    M \circ \phi =\sum_{q \in \mathcal{Z}(\phi(p))} M(q) \prod_{d \in\{x, y, z\}}\left(1-\left|\phi_d(p)-q_d\right|\right) \,.
\end{equation}
Here $ \mathcal{Z}(\phi(p))$ are the voxel neighbors of $\phi(p)$.

 \section{Experiment and Results}

 To evaluate the proposed method and its variants under diverse conditions, we considered three scenarios using structural volumes acquired with hydrogen and phosphorus head coils: (1) with the skull included, (2) without the skull for rigid registration, and (3) with the skull removed for deformable registration. We compared four variants of our method against established approaches, including SyN \cite{Avants2008}, VoxelMorph \cite{Balakrishnan2019}, ViT-V-Net \cite{chen2021vit}, and TransMorph \cite{chen2022transmorph}. For rigid registration, we adapted the outputs of VoxelMorph and ViT-V-Net to predict affine parameters, while the affine configuration of TransMorph was used as described in its original publication.

All evaluations were conducted on the dataset provided by the Medical University of Innsbruck (Ethical approval number: UN5100, Sitzungsnummer: 325/4.19), which comprises 60 pairs of intra-patient T1- and T2-weighted MR images acquired on a 3T Siemens scanner. Imaging was performed using a 64-channel ${}^{1}\text{H}$ head-neck coil (Siemens, Erlangen, Germany) and a double-tuned ${}^{1}\text{H}$/${}^{31}\text{P}$ volume head coil (Rapid Biomedical, Würzburg, Germany). From this dataset, 30 pairs were used for training, 10 for validation, and 20 for testing. These multimodal acquisitions enabled robust assessment of our registration method across different coil types and structural contrasts. Additionally, we conducted further evaluations on two public datasets: IXI (atlas-to-patient brain MRI) and OASIS (inter-patient brain MRI), to validate the generalizability of our approach.

In our approach to estimating uncertainty, we utilize the Monte Carlo Dropout method. This technique involves applying dropout both during the training phase and at inference time, effectively generating an ensemble of models by selectively omitting neurons from the network at random. To assess uncertainty, we perform several forward passes through the network with dropout active, each time with a different random configuration of neuron activations. Specifically, we process each input through the network 10 times, with a unique dropout mask applied each time. This results in 10 distinct predictions per input, from which we can compute the mean and variance. The variance among these predictions serves as an indicator of the model's uncertainty. By averaging the outcomes of these multiple passes, we obtain a more stable prediction, while the variance among them provides a measure of the uncertainty tied to the prediction. This method allows us to pinpoint areas where the model's confidence is lower, offering critical insights into the trustworthiness of the predictions.
 \subsection{Rigid registration with skull intact}
\begin{figure}[thb!]
\centering
\includegraphics[width=1\linewidth]{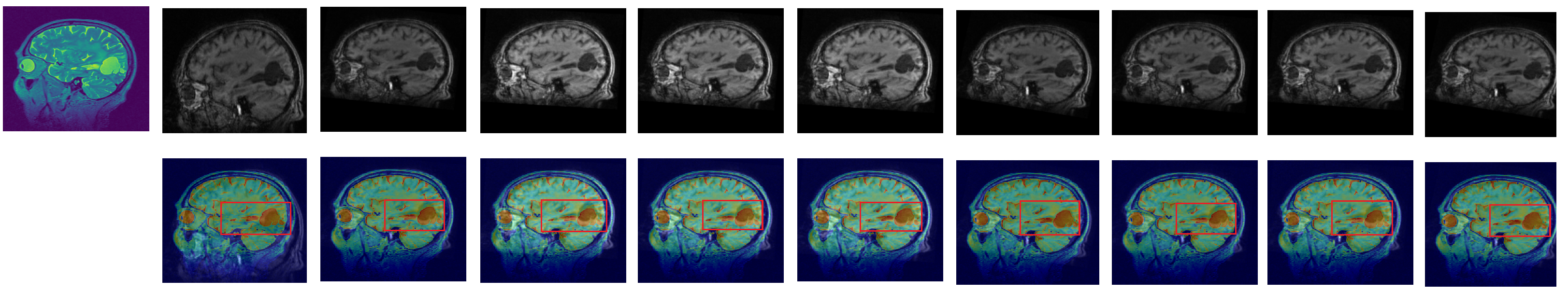}
\caption{(left to right) ${}^{31}\text{P}$ structural volume, ${}^{1}\text{H}$ structural volume, ANTs, VoxelMorph, ViT-V-Net, TransMorph, Reg-LEdge-Model Variant-1,Reg-LEdge-Model Variant-2,Reg-LEdge-Model Variant-3,Reg-LEdge-Model Variant-4. Bottom row shows the residual images $( f - m \circ \phi)$}
\label{fig:rigid_with_skull}
\end{figure}

\begin{figure}[thb!]
\centering
\includegraphics[width=0.7\linewidth]{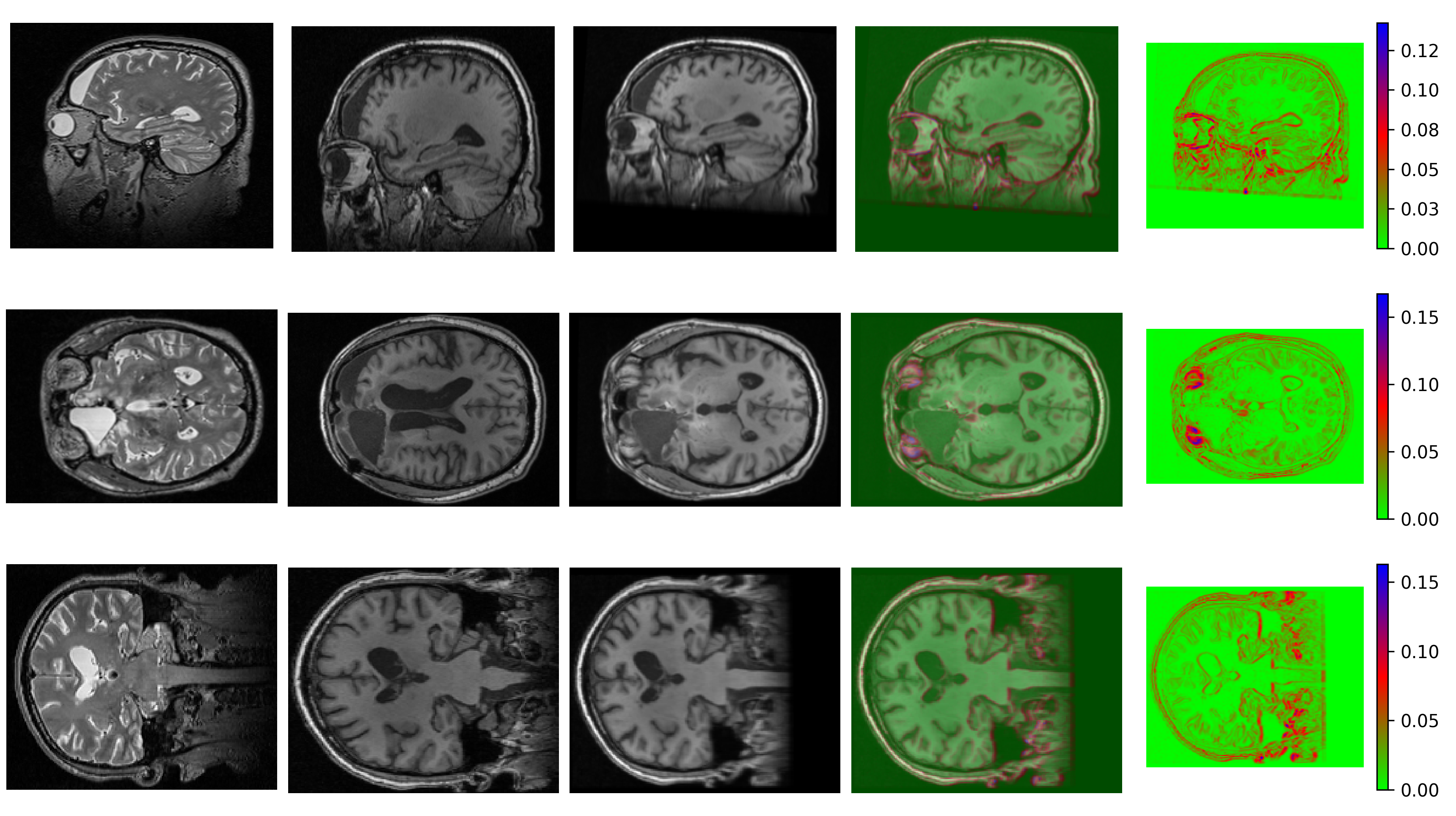}
\caption{(Left to right) Moving image, fixed image, mean prediction (10 passes), uncertainty overlay and uncertainty maps. }
\label{fig:MC_rigid_with_skull}
\end{figure}

 The results presented in the table \ref{tab:rigid_skull} illustrate the performance of various methods in terms of Dice similarity coefficients for WM-GM (white-gray matter) and brain mask for rigid registration task without skull stripping. Among the methods compared, ours Reg-LEdge-Model variant-4 demonstrates the highest performance with a Dice WM-GM score of 0.784 however, ANTs has better performed in Dice score of brain mask 0.924 compared to our variant-4 0.917. This suggests superior registration accuracy in white-gray matter and comparable in brain mask delineations. Reg-LEdge-Model variant-3 follows closely, achieving a Dice WM-GM of 0.763 and a Dice brain mask of 0.881. Other variants of our method also perform well, consistently surpassing the benchmarks set by VM-affine, TransMorph-affine, and ViT-V-Net. Notably, TransMorph-affine shows competitive performance with a Dice brain mask score of 0.873, yet falls short in WM-GM Dice compared to our methods. These results highlight the efficacy of our proposed methodologies in enhancing rigid registration accuracy, which is crucial for precise neuro-imaging analysis. These results endorse that the proposed learnable edge detection module helps in aligning anatomical structures better than other methods. This can also be seen in qualitative results shown in Figure \ref{fig:rigid_with_skull} and the uncertainty estimation of the Reg-LEdge-Model variant-4 can be seen in Figure \ref{fig:MC_rigid_with_skull}.

 \begin{table}[thb!]
\caption{Comparison of Methods on volumes with skull intact}
\begin{center}
\begin{tabular}{|c|c|c|}
\hline
Method&  Dice (WM-GM)& Dice (brain mask)\\
\hline
ANTs & \(0.742 \pm 0.122\) & \(0.924 \pm 0.136\) \\
\hline
VoxelMorph-affine & \(0.732 \pm 0.133\) & \(0.834 \pm 0.121\) \\
\hline
TransMorph-affine & \(0.745 \pm 0.213\) & \(0.873 \pm 0.112\)  \\
\hline
ViT-V-Net-affine & \(0.723 \pm 0.142\) & \(0.861 \pm 0.121\) \\
\hline
Reg-LEdge-Model Variant-1 & \(0.746 \pm 0.123\) & \(0.836 \pm 0.112\)  \\
\hline
Reg-LEdge-Model Variant-2  & \(0.759 \pm 0.106\) & \(0.872 \pm 0.155\)  \\
\hline
Reg-LEdge-Model Variant-3  & \(0.763 \pm 0.186\) & \(0.881 \pm 0.107\)  \\
\hline
Reg-LEdge-Model Variant-4  & \(0.784 \pm 0.125\) & \(0.917 \pm 0.101\)  \\
\hline
\end{tabular}
\label{tab:rigid_skull}
\end{center}
\end{table}

 \subsection{Rigid registration with skull removed}
\begin{figure}[thb!]
\centering
\includegraphics[width=1\linewidth]{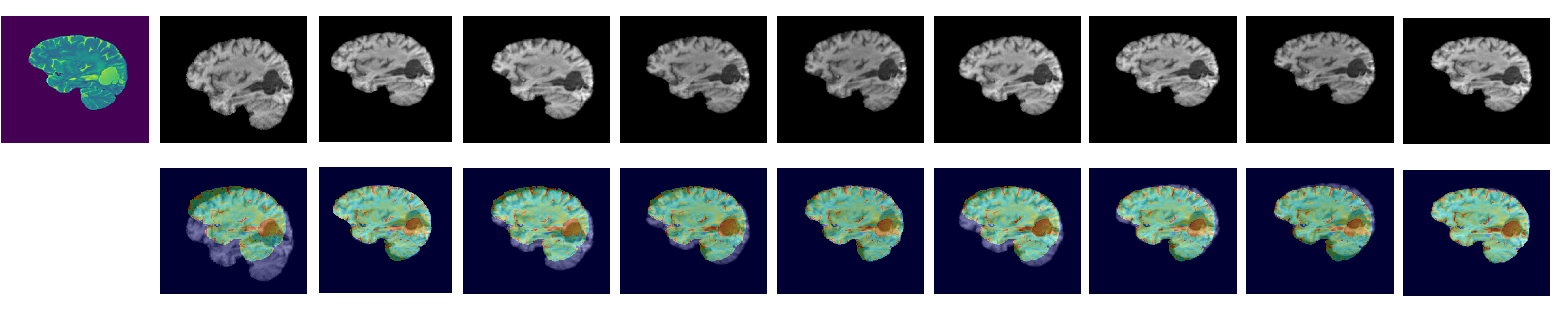}
\caption{(left to right) ${}^{31}\text{P}$ structural volume, ${}^{1}\text{H}$ structural volume, ANTs, VoxelMorph, ViT-V-Net, TransMorph, Reg-LEdge-Model Variant-1,Reg-LEdge-Model Variant-2, Reg-LEdge-Model Variant-3, Reg-LEdge-Model Variant-4. Bottom row shows the residual images $( f - m \circ \phi)$}
\label{fig:rigid_without_skull}
\end{figure}

\begin{figure}[thb!]
\centering
\includegraphics[width=0.7\linewidth]{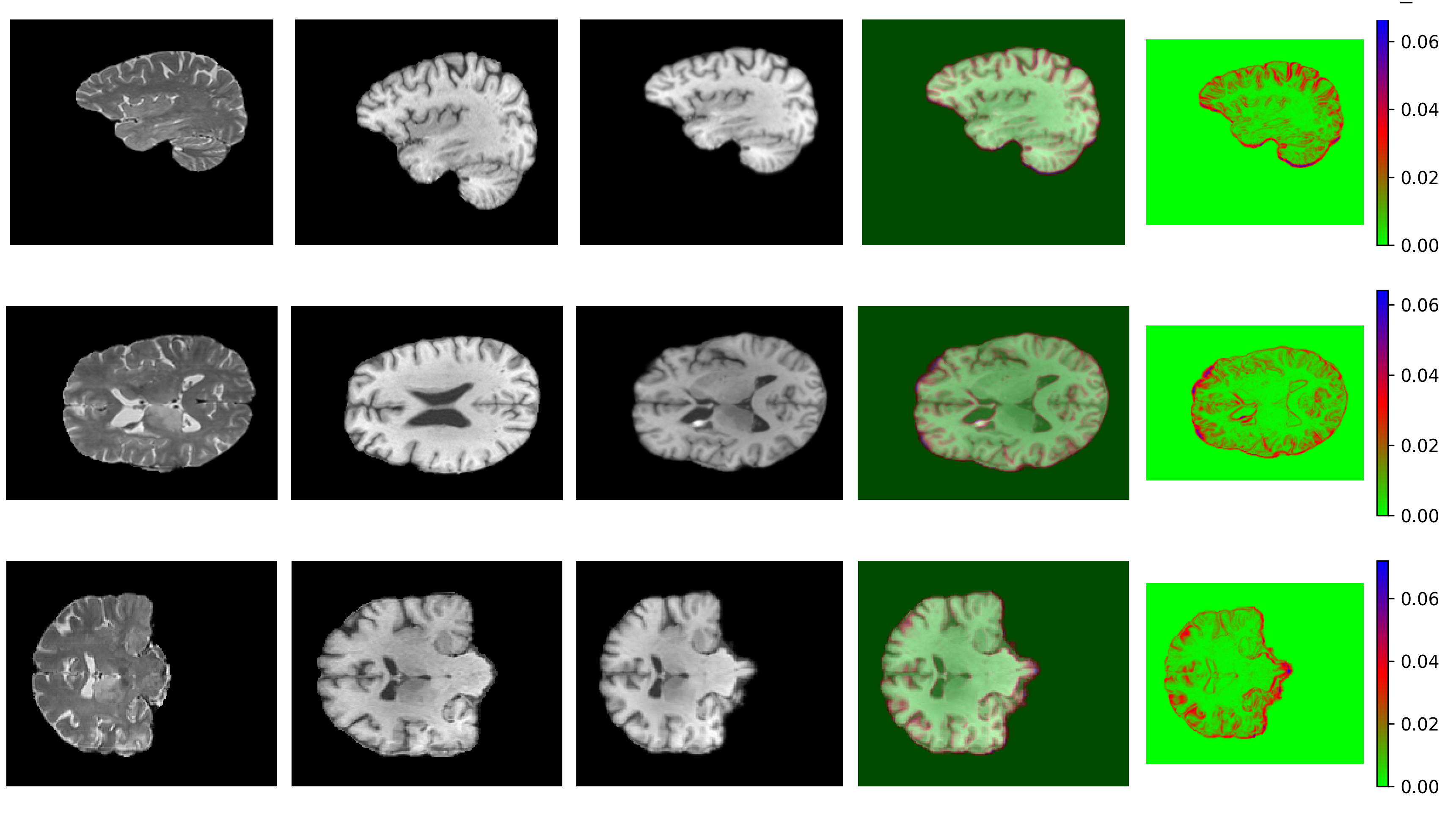}
\caption{(Left to right) Moving image, fixed image, mean prediction (10 passes), uncertainty overlay and uncertainty maps.}
\label{fig:MC_rigid_without_skull}
\end{figure}

 The table \ref{tab:rigid_without_skull} illustrates the performance of different methods in terms of Dice similarity coefficients for WM-GM (white-gray matter) and brain mask for rigid registration tasks with skull stripping. Among the methods evaluated, ours Reg-LEdge-Model variant-4 achieves the highest scores, with a Dice WM-GM of 0.757 and a Dice brain mask of 0.912, indicating superior registration accuracy. On the other hand ANTs also demonstrate strong performance, particularly in the brain mask 0.911 however, Dice WM-GM 0.726 is not comparable with our method Reg-LEdge-Model variant-4. Reg-LEdge-Model variant-2 and Reg-LEdge-Model variant-3 perform better, particularly in the brain mask task, with Dice scores of 0.851 and 0.845, respectively, though their WM-GM scores are slightly lower at 0.735 and 0.734.The methods VM-affine, TransMorph-affine, show competitive but lower performance compared to our methods, with Dice WM-GM scores ranging from 0.724 to 0.731 and Dice brain mask scores between 0.792 and 0.812. Reg-LEdge-Model variant-1 achieves a Dice WM-GM score of 0.741 and a Dice Brain Mask score of 0.803, indicating a balanced performance but not as high as the main Reg-LEdge-Model variant-4 method. Figure \ref{fig:rigid_without_skull} shows the visual comparison of different methods and Figure \ref{fig:MC_rigid_without_skull} shows uncertainty estimation.

 \begin{table}[thb!]
\caption{Comparison of Methods on volumes with skull removed}
\begin{center}
\begin{tabular}{|c|c|c|}
\hline
Method&  Dice (WM-GM)& Dice (brain mask)\\
\hline
ANTs & \(0.726 \pm 0.122\) & \(0.911 \pm 0.136\) \\
\hline
VoxelMorph-affine & \(0.724 \pm 0.163\) & \(0.792 \pm 0.121\) \\
\hline
TransMorph-affine & \(0.731 \pm 0.111\) & \(0.812 \pm 0.138\)  \\
\hline
ViT-V-Net-affine & \(0.729 \pm 0.112\) & \(0.804 \pm 0.141\) \\
\hline
Reg-LEdge-Model Variant-1 & \(0.741 \pm 0.123\) & \(0.803 \pm 0.152\)  \\
\hline
Reg-LEdge-Model Variant-2  & \(0.735 \pm 0.116\) & \(0.851 \pm 0.126\)  \\
\hline
Reg-LEdge-Model Variant-3  & \(0.734 \pm 0.118\) & \(0.845 \pm 0.124\)  \\
\hline
Reg-LEdge-Model Variant-4  & \(0.757 \pm 0.116\) & \(0.912 \pm 0.161\)  \\
\hline
\end{tabular}
\label{tab:rigid_without_skull}
\end{center}
\end{table}

 \begin{figure}[thb!]
\centering
\includegraphics[width=0.7\linewidth]{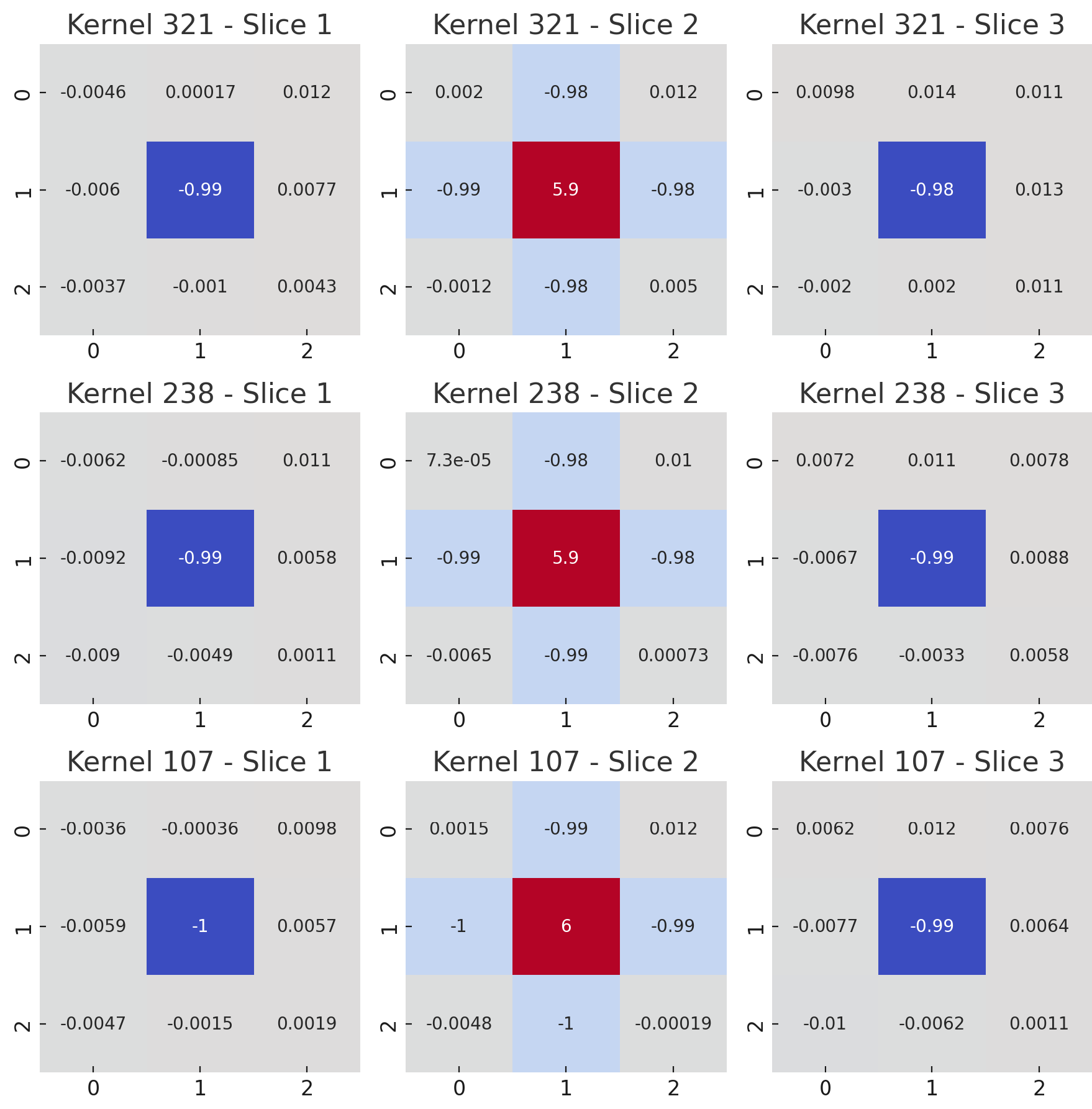}
\caption{Visualization of randomly selected 3D edge detection kernels, showing their spatial structures across different depth slices.}
\label{fig:kernals}
\end{figure}

In Figure \ref{fig:kernals} Kernel Heatmaps visualize randomly selected edge detection filters, showing their spatial structure across three depth slices. Each row corresponds to a different kernel, while each column represents a slice from the 3D kernel along the depth axis. The color intensity in each heatmap indicates the weight values assigned by the filter, where positive values enhance certain features and negative values suppress or detect edges. If the center pixel has a high positive value, it means the kernel is enhancing central features, while surrounding pixels with negative values indicate contrast edge detection. The differences in patterns among kernels determine how each filter responds to the same input. The center values often have higher magnitudes, signifying strong edge detection properties, whereas surrounding values influence how the filter interacts with neighboring pixels. These kernels play a crucial role in 3D feature extraction by identifying spatial patterns in volumetric data, particularly in medical imaging and segmentation tasks.

\begin{figure}[thb!]
\centering
\includegraphics[width=0.9\linewidth]{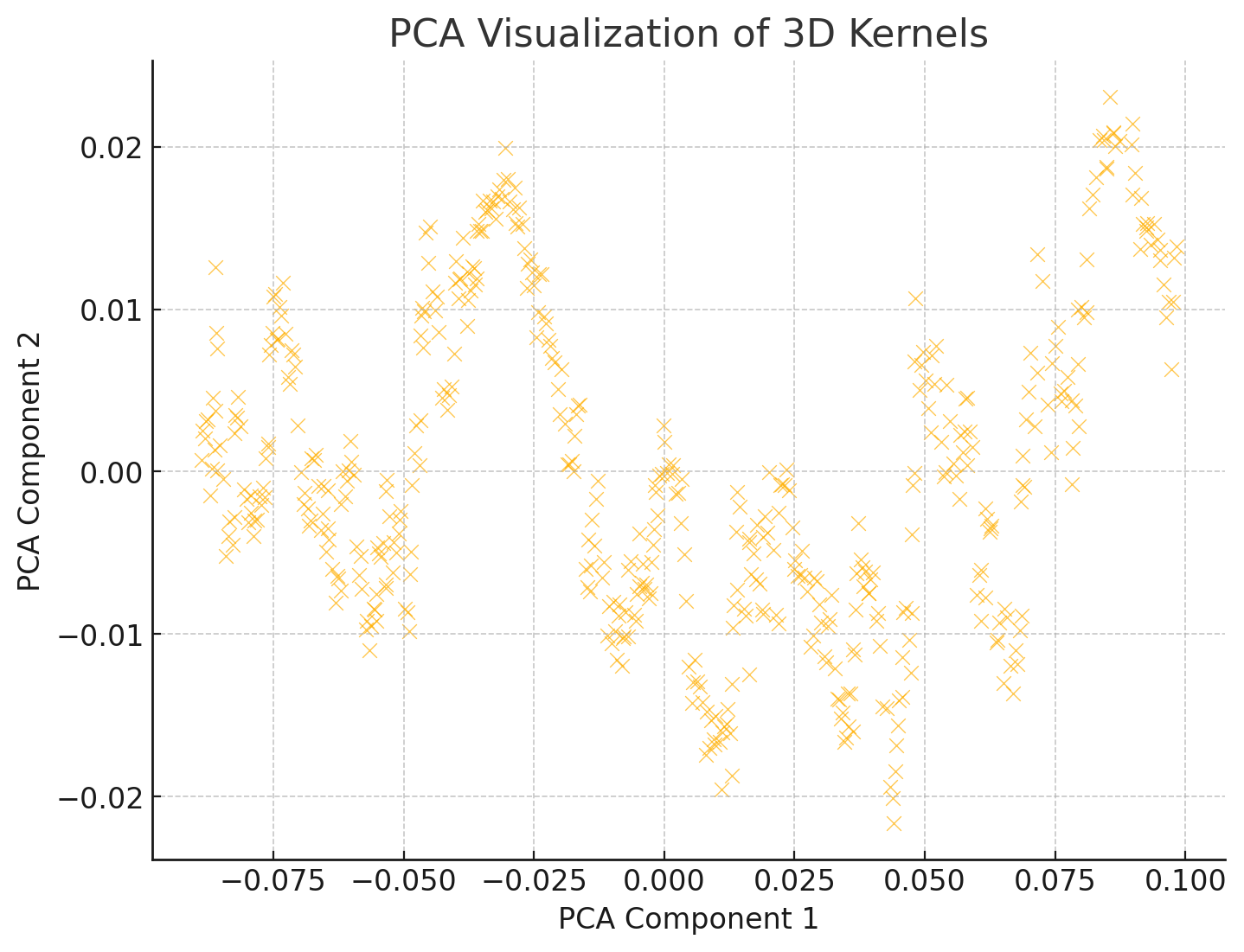}
\caption{PCA scatter plot of 500 3D edge detection kernels, showing their distribution in a reduced 2D feature space based on principal component analysis}
\label{fig:PCA}
\end{figure}

The PCA Scatter Plot in Figure \ref{fig:PCA} represents the dimensionality reduction of the 3D edge detection kernels, reducing their original 27-dimensional representation (from \(3 \times 3 \times 3\)) into a 2D space for visualization. Each point in the plot corresponds to a kernel, with its position determined by the first two principal components that capture the most variance in the data. Kernels that are closer together share similar structures and filtering properties, while those farther apart represent more distinct features. This visualization helps assess the diversity of the learned kernels, indicating whether they cover a broad range of edge-detecting patterns or are highly similar. The distribution suggests that while many kernels exhibit similarities (close together in the plot), there is also notable diversity (points further apart), indicating that the kernels capture a range of edge-detection and spatial filtering properties.

\begin{figure}[thb!]
\centering
\includegraphics[width=1\linewidth]{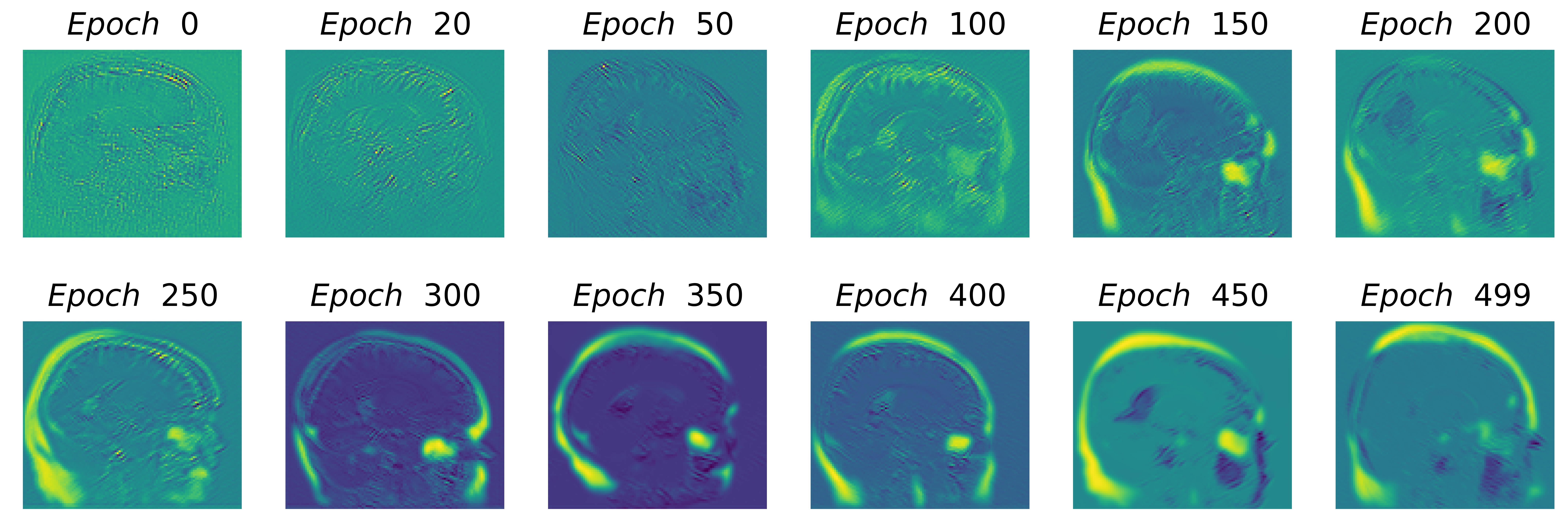}
\caption{Evolution of 3D edge detection kernels across 500 epochs, showing an initial focus on brain features and a later shift towards emphasizing skull edges for precise image registration}
\label{fig:K_eval}
\end{figure}

Figure \ref{fig:K_eval} illustrates the evolution of the kernel's learning process across 500 epochs. Early in training, the kernels focus primarily on extracting general brain features, as observed in the initial epochs where the features are broad and dispersed. As the training progresses, particularly beyond epoch 150, the kernels adapt to emphasize specific skull edges and boundaries, aligning with the task of image registration. This transition reflects how the model gradually shifts its focus from coarse to finer features, eventually fine-tuning its detection towards anatomical structures critical for registration accuracy.

\subsection{Non-rigid registration}
\begin{figure}[thb!]
\centering
\includegraphics[width=1\linewidth]{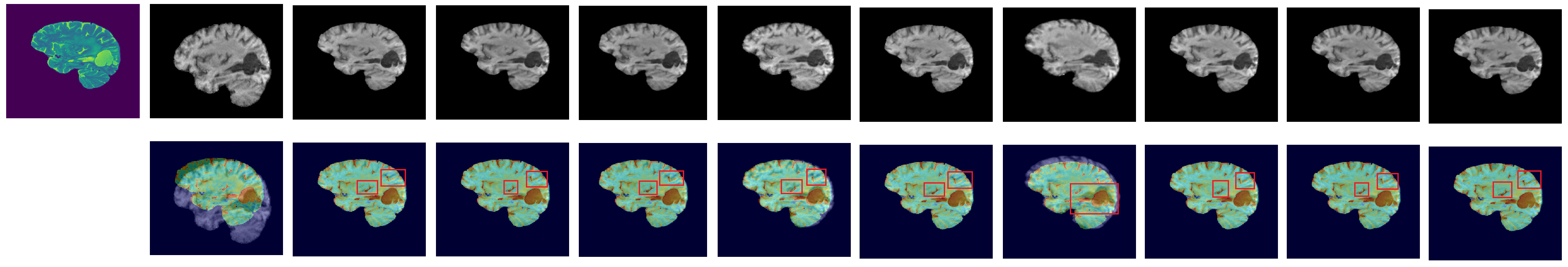}
\caption{(left to right) ${}^{31}\text{P}$ structural volume, ${}^{1}\text{H}$ structural volume,SyN, LDDMM, VoxelMorph, ViT-V-Net, TransMorph, LEdge-U-Model Variant-1, LEdge-U-Model Variant-2, LEdge-U-Model Variant-3, LEdge-U-Model Variant-4. Bottom row shows the residual images $( f - m \circ \phi)$}
\label{fig:non-rigid}
\end{figure}

\begin{figure}[thb!]
\centering
\includegraphics[width=0.7\linewidth]{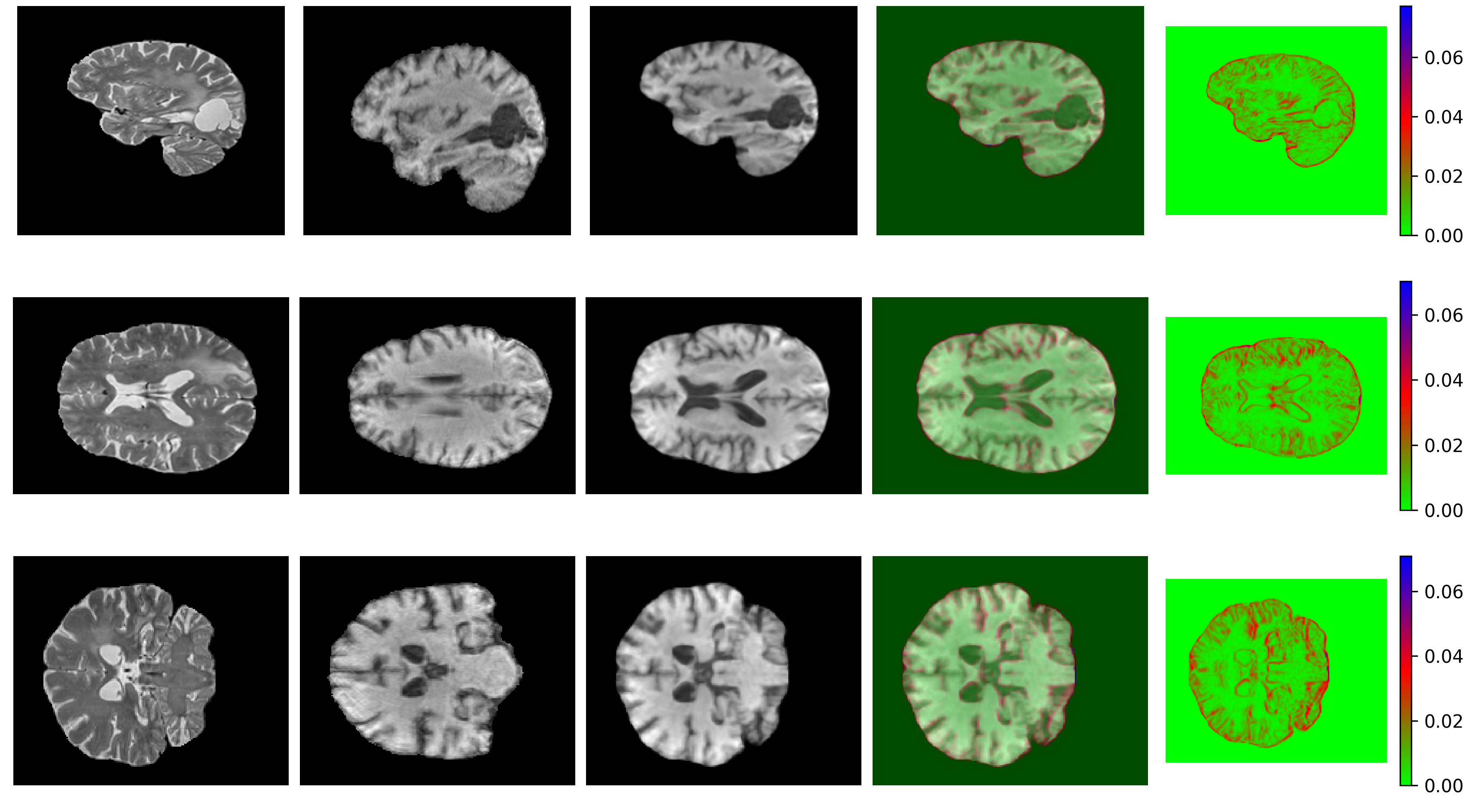}
\caption{(Left to right) Moving image, fixed image, mean prediction (10 passes), uncertainty overlay and uncertainty maps.}
\label{fig:MC_non_rigid_with_noskull}
\end{figure}

For elastic registration, we initially used our method Reg-LEdge-Model variant-4 (rigid registration) to perform affine registration on all images. These affine-registered images were then used as input for our proposed Reg-LEdge-U-Model variants of non-rigid registration, allowing us to compare their performance with other methods.
The table \ref{tab:non_rigid} presents the comparison of various registration methods based on their Dice similarity coefficients for WM-GM (white-gray matter) segmentation. Among the methods evaluated, LEdge-U-Model variant-4 achieves the highest performance with a Dice WM-GM score of 0.798, indicating the most accurate registration. Following closely are LEdge-U-Model variant-3 and LEdge-U-Model variant-2 with Dice scores of 0.791 and 0.783, respectively, showcasing the effectiveness of our proposed variants in improving segmentation accuracy.
LEdge-U-Model variant-1 also performs well, with a Dice score of 0.774, slightly surpassing TransMorph which has a score of 0.772. VoxelMorph show competitive performance with Dice scores of 0.769 and 0.761, respectively. ViT-V-Net  on the other hand, exhibit lower accuracy with Dice scores of 0.746. SyN and LDDMM secure Dice score of 0.766 and 0.754 respectively.These results underscore the superiority of our proposed methods in enhancing registration accuracy, as reflected by their higher Dice WM-GM scores compared to other methods in the table. Figure \ref{fig:non-rigid} shows qualitative results and Figure \ref{fig:MC_non_rigid_with_noskull} represents the uncertainty estimation of the best performing proposed Reg-LEdge-U-Model variant-4.

 \begin{table}[thb!]
\caption{Comparison of Methods for non-rigid registration}
\begin{center}
\begin{tabular}{|c|c|c|c|}
\hline
Method&  Dice (WM-GM)& $\% \, of \, |J_\phi| \le 0$\\
\hline
SyN & \(0.766 \pm 0.136\) & $<0.0001$  \\
\hline
LDDMM & \(0.754 \pm 0.186\) & $<0.0001$  \\
\hline
VoxelMorph & \(0.769 \pm 0.163\) & $0.0006$ \\
\hline
TransMorph & \(0.772 \pm 0.111\) & $0.0008$  \\
\hline
ViT-V-Net & \(0.746 \pm 0.112\) & $0.0008$ \\
\hline
LEdge-U-Model Variant-1 & \(0.774 \pm 0.123\) & $1.7621$  \\
\hline
LEdge-U-Model Variant-2  & \(0.783 \pm 0.116\) & $0.0002$  \\
\hline
LEdge-U-Model Variant-3  & \(0.791 \pm 0.118\) & $<0.0001$ \\
\hline
LEdge-U-Model Variant-4  & \(0.798 \pm 0.116\) & $<0.0001$ \\
\hline
\end{tabular}
\label{tab:non_rigid}
\end{center}
\end{table}

\subsection{Additional evaluation}
We have conducted additional evaluation of proposed non-rigid registration variants on two datasets. For atlas-to-patient brain MRI registration, we use the IXI dataset. we utilized the preprocessed IXI dataset from \cite{chen2022transmorph}, we train on 40 samples, validate on 20, and test on 115 samples, with each volume resized to $160 \times 192 \times 224$ pixels, and performance evaluated using 29 anatomical structure labels. For inter-patient brain MRI registration, we use the OASIS dataset from the 2021 Learn2Reg challenge, preprocessed by \cite{chen2022transmorph}. This involves 25 training samples and 19 validation samples, with each image skull-stripped, aligned, and normalized to a resolution of 160 × 192 × 224 pixels. Registration performance is measured using the Dice Score on 29 anatomical structures. Our method's results on the atlas-to-patient dataset are compared with VoxelMorph-1, VoxelMorph-2, ViT-V-Net, nn-Former, CoTr, and TransMorph, following the parameters in \cite{chen2022transmorph}, focusing on the average Dice score on validation, Dice score on test samples, and the fraction of voxels with non-positive Jacobian determinants (FNJ).

\begin{table}[thb!]
\centering
\begin{tabular}{ |l||l|l|l|l|l|}
 \toprule
 \multicolumn{4}{|c|}{Atlas-to-patient MRI dataset} \\
 \midrule
Method& Val.  Dice& Test Dice& $\% \, of \, |J_\phi| \le 0$\\
 \midrule
SyN& -& 0.624$\pm$ 0.141& $<$ 0.001\\
LDDMM& -& 0.662$\pm$ 0.164& $<$ 0.001\\
VoxelMorph-1& 0.701$\pm$ 0.134& 0.701 $\pm$ 0.138& $<$ 0.001\\
VoxelMorph-2& 0.709 $\pm$ 0.105& 0.704 $\pm$ 0.152& $<$ 0.001\\
CycleMorph& 0.711$\pm$ 0.108& 0.705$\pm$ 0.132& 0.012 $<$ 0.001\\
ViT-V-Net& 0.671 $\pm$ 0.104& 0.662 $\pm$ 0.181& 0.018 $<$ 0.001\\
nnFormer& 0.711 $\pm$ 0.102& 0.701 $\pm$ 0.124& 0.011 $<$ 0.001\\
CoTr& 0.613 $\pm$ 0.118& 0.602 $\pm$ 0.115& 0.003 $<$ 0.001\\
TransMorph& 0.719 $\pm$ 0.101& 0.701 $\pm$ 0.064&$<$ 0.001\\
Reg-LEdge-U-Model variant-1& 0.723 $\pm$0.121& 0.723 $\pm$ 0.122& 0.025 $<$ 0.001\\
Reg-LEdge-U-Model variant-2& 0.610 $\pm$0.144& 0.612 $\pm$ 0.146& 0.038 $<$ 0.001\\
Reg-LEdge-U-Model variant-3& 0.728 $\pm$0.110& 0.726 $\pm$ 0.139& $<$ 0.001\\
Reg-LEdge-U-Model variant-4& 0.721 $\pm$0.118& 0.713 $\pm$ 0.142& $<$ 0.001\\
 \bottomrule
\end{tabular}
\caption{Quantitative results of different models on Atlas-to-patient MRI dataset}
\label{table:1}
\end{table}

\begin{table}[thb!]
\centering
\begin{tabular}{ |l||l|l|l|}
 \toprule
 \multicolumn{4}{|c|}{Inter-patient MRI dataset} \\
 \midrule
 Method& Val. Dice& Test Dice&  $\% \, of \,  |J_\phi| \le 0$\\
 \midrule
 SyN& -& 0.746$\pm$ 0.156& $<$ 0.001\\
LDDMM& -& 0.741$\pm$ 0.124& $<$ 0.001\\
VoxelMorph-1& 0.742 $\pm$ 0.191& 0.738$\pm$0.181& $<$ 0.001\\
VoxelMorph-2& 0.752$\pm$0.126& 0.749$\pm$0.112& $<$ 0.001\\
CycleMorph& 0.761$\pm$ 0.182& 0.758$\pm$ 0.122& 0.112 $<$ 0.001\\
ViT-V-Net&  0.781$\pm$0.121& 0.773$\pm$0.110& 0.008$<$ 0.001\\
nnFormer& 0.774$\pm$0.114& 0.779$\pm$0.122& $<$ 0.001 \\
CoTr&  0.702$\pm$0.116& 0.696$\pm$0.024& 0.106$<$ 0.001 \\
TransMorph& 0.785$\pm$0.101& 0.780$\pm$0.124& $<$ 0.001\\
Reg-LEdge-U-Model variant-1& 0.763 $\pm$0.141& 0.753 $\pm$ 0.122& 0.045 $<$ 0.001\\
Reg-LEdge-U-Model variant-2& 0.782 $\pm$0.164& 0.782 $\pm$ 0.146& 0.066 $<$ 0.001\\
Reg-LEdge-U-Model variant-3& 0.804 $\pm$0.135& 0.796 $\pm$ 0.139& $<$ 0.001\\
Reg-LEdge-U-Model variant-4& 0.811 $\pm$0.129& 0.803 $\pm$ 0.142&$<$ 0.001\\
 \bottomrule
\end{tabular}

\caption{Quantitative results of different models on Inter-patient MRI dataset}
\label{table:2}
\end{table}

\section*{Ethics Approval Statement}

This study includes data provided by the Medical University of Innsbruck. Ethical approval for the use of this data was granted by the institutional review board under approval number: UN5100, Sitzungsnummer: 325/4.19.

\section*{Acknowledgment}

This work was supported by the Austrian Science Fund (FWF) [grant number DOC 110]
\section{Conclusion}
The proposed registration method, which combines learning-based rigid and non-rigid registration with learnable edge modules, has demonstrated superior performance across different registration setups. This approach effectively aligns key anatomical structures, addressing challenges in both rigid and non-rigid registration by enhancing feature recognition and accommodating variations in orientation, scale, and tissue deformation. The results, validated on the Medical University of Innsbruck dataset along with two publicly available datasets highlight the method's ability to outperform state-of-the-art techniques, thus improving the accuracy and quality of mono and multi-modal imaging analyses. Ultimately, this enhanced registration framework contributes to more precise preoperative assessments, better-informed treatment planning, and improved patient outcomes in neuro-oncology.
 \bibliographystyle{elsarticle-num}
\bibliography{references}
\end{document}